\documentclass[10pt,a4paper,twocolumn,fleqn]{article}
\usepackage[T1]{fontenc}
\usepackage[utf8]{inputenc}
\usepackage[a4paper,left=17.5mm,right=17.5mm,top=20mm,bottom=20mm,columnsep=18pt]{geometry}
\usepackage{amsmath}
\usepackage{newtxtext,newtxmath}
\usepackage{bm}
\usepackage{graphicx,makecell,booktabs,tabularx,multirow,float}
\usepackage{algorithm,algpseudocode}
\usepackage{caption}
\usepackage{titlesec}
\usepackage{microtype}
\usepackage[numbers,super,sort&compress]{natbib}
\usepackage{xcolor}
\definecolor{linkblue}{RGB}{25,65,115}
\usepackage[colorlinks=true,allcolors=linkblue]{hyperref}
\usepackage[capitalize,nameinlink]{cleveref}
\hypersetup{
  pdftitle={A physics-constrained deep reinforcement learning method for autonomous mission planning of Titan deep space probes},
  pdfauthor={Hanmin Li; Tao Wang; Shuangliang Liu; Zhe Zhang; Maodeng Li},
  pdfsubject={Author preprint},
  pdfkeywords={Titan deep space probe, mission planning, uncertain systems, deep reinforcement learning, Markov decision process}
}
\newcommand{\figref}[1]{\hyperref[#1]{Fig.~\ref*{#1}}}
\newcommand{\tabref}[1]{\hyperref[#1]{Table~\ref*{#1}}}
\newcommand{\secref}[1]{\hyperref[#1]{Section~\ref*{#1}}}
\makeatletter
\renewcommand\normalsize{%
  \@setfontsize\normalsize{9.5}{11.4}%
  \abovedisplayskip 9pt plus 2pt minus 4pt
  \abovedisplayshortskip 0pt plus 3pt
  \belowdisplayshortskip 5pt plus 3pt minus 3pt
  \belowdisplayskip\abovedisplayskip
  \let\@listi\@listI}
\renewcommand\small{%
  \@setfontsize\small{8.5}{10.2}%
  \abovedisplayskip 8.5pt plus 3pt minus 4pt
  \abovedisplayshortskip 0pt plus 2pt
  \belowdisplayshortskip 4pt plus 2pt minus 2pt
  \belowdisplayskip\abovedisplayskip}
\makeatother
\normalsize
\titleformat{\section}{\normalfont\normalsize\bfseries}{\thesection.}{0.5em}{}
\titleformat{\subsection}{\normalfont\normalsize\itshape}{\thesubsection.}{0.5em}{}
\titleformat{\subsubsection}{\normalfont\normalsize\itshape}{\thesubsubsection.}{0.5em}{}
\titlespacing*{\section}{0pt}{14pt plus 3pt minus 2pt}{7pt plus 2pt minus 1pt}
\titlespacing*{\subsection}{0pt}{11pt plus 3pt minus 2pt}{4pt plus 2pt minus 1pt}
\titlespacing*{\subsubsection}{0pt}{9pt plus 2pt minus 1pt}{3pt}

\DeclareCaptionFont{preprintcaption}{\fontsize{8.5}{10.2}\selectfont}
\begin{document}
\twocolumn[{
\begin{center}
{\fontsize{18}{21.5}\selectfont\bfseries
A physics-constrained deep reinforcement learning method for autonomous mission planning of Titan deep space probes\par}
\vspace{10pt}
{\fontsize{11}{14}\selectfont
Hanmin Li\textsuperscript{a},
Tao Wang\textsuperscript{a,*},
Shuangliang Liu\textsuperscript{b},
Zhe Zhang\textsuperscript{b},
Maodeng Li\textsuperscript{b}\par}
\vspace{6pt}
{\fontsize{8.5}{10.5}\selectfont
\textsuperscript{a}School of Intelligent Systems Engineering, Sun Yat-sen University, Shenzhen, 518107, China\par
\textsuperscript{b}Deep Space Exploration Laboratory, Hefei, 230031, China\par
\vspace{3pt}
\textsuperscript{*}Corresponding author: Tao Wang. Email: \href{mailto:wangt339@mail.sysu.edu.cn}{wangt339@mail.sysu.edu.cn}\par}
\end{center}
\vspace{6pt}
\begin{center}
\begin{minipage}{0.94\textwidth}
\fontsize{9}{11}\selectfont
\noindent\textbf{Abstract}\par\vspace{3pt}
\noindent
Titan is a compelling target for future deep space exploration, while its extreme Earth-Titan distance and complex environment make continuous ground-in-the-loop planning impractical. Existing planning methods, primarily developed for Earth-observation scenarios, have limited generalizability to uncertain deep space conditions and adaptability to time-varying mission states. To address these challenges, this paper proposes a physics-constrained maskable proximal policy optimization (PCM-PPO) method for Titan autonomous mission planning. A Titan-specific discrete-time Markov decision process is established to characterize mission feasibility and resource constraints. A system dynamics and geometric model is incorporated to enhance physical representation, including central gravity, J2 perturbation, third-body attraction, and downlink occultation. To accommodate opportunity targets under a bounded action dimension, PCM-PPO organizes predefined and opportunity targets through a fixed-capacity target pool and represents the mission state with a target-wise state matrix and a global state vector. A physics-constrained action mask constructs an admissible decision domain from current mission states, guiding policy learning toward physically executable decisions and avoiding infeasible exploration. An integrated training strategy facilitates policy optimization under coupled constraints. Comparative, ablation, and uncertainty experiments validate the effectiveness of PCM-PPO. PCM-PPO provides a feasible and interpretable scheme for autonomous mission planning of Titan deep space probes.
\par\vspace{6pt}
\noindent\textbf{Keywords:} Titan deep space probe; Mission planning; Uncertain systems; Deep reinforcement learning; Markov decision process
\end{minipage}
\end{center}
\vspace{9pt}
}]
\section{Introduction}
\label{1}

Deep space exploration is rapidly advancing toward increasingly distant and complex planetary environments, among which Titan, Saturn’s largest moon, has become a major target of scientific interest \cite{horst2017titan}. Owing to its dense nitrogen-rich atmosphere, active hydrocarbon cycle, and potential prebiotic chemical environment, Titan is considered one of the most compelling bodies for astrobiological investigation in the solar system \cite{hayes2016lakes,raulin2012prebiotic}. Its scientific significance has been highlighted by the Cassini-Huygens mission \cite{spilker2019cassini} and is further reinforced by the ongoing development of NASA’s Dragonfly rotorcraft mission \cite{barnes2021science}. However, the extreme Earth–Titan distance leads to substantial communication delays, making continuous ground-in-the-loop mission management increasingly impractical \cite{nixon2018titan}. Therefore, future Titan deep space probes require advanced onboard autonomous mission planning capabilities to generate and adapt observation schedules with limited human intervention.

Deep space mission planning is inherently an NP-hard optimization problem \cite{lemaitre2002selecting,qi2024mission,junhui2026relaxed}, since numerous observation tasks must be coordinated under limited time windows, onboard resources, and multiple constraints \cite{zhang2024trajectory,gu2022large}. A large number of researchers have proposed various optimization methods, including heuristic, metaheuristic, and multi-objective evolutionary approaches \cite{qi2024cooperative}. Heuristic methods typically improve planning efficiency by exploiting problem-specific structures or scheduling rules \cite{song2022two,chang2023solving,wang2018exact}. For example, Rigo et al. \cite{rigo2022branch} developed a branch-and-price algorithm for nanosatellite task scheduling by combining Dantzig–Wolfe decomposition with dynamic programming-based column generation, thereby improving mission quality of service. Chien et al. \cite{chien2015heuristic} proposed a greedy constructive onboard rescheduling strategy for an Earth-observing spacecraft, which enables rapid response through model-based constraint handling. Metaheuristic algorithms have also been widely adopted to enhance global search capability in large-scale combinatorial planning problems \cite{shang2025joint}. Liu et al. \cite{liu2017adaptive} proposed an adaptive large neighborhood search method for agile satellite scheduling with time-dependent transition times. Similarly, Han et al. \cite{han2022simulated} designed an improved simulated-annealing-based heuristic to address multiple agile satellite scheduling under cloud coverage uncertainty. Zhang et al. \cite{zhang2022improved} further formulated an integrated satellite imaging and data transmission scheduling problem and solved it using an improved genetic algorithm. For scenarios involving multiple conflicting objectives, multi-objective evolutionary methods have been introduced \cite{wu2022ensemble}. Tangpattanakul et al. \cite{tangpattanakul2015multi} developed an indicator-based multi-objective local search heuristic for agile satellite observation scheduling, and Du et al. \cite{du2019moea} proposed MOEA-based memetic algorithms for satellite range scheduling to simultaneously optimize competing scheduling objectives.
However, autonomous mission planning for Titan exploration is considerably more challenging than conventional satellite scheduling problems. A Titan deep space probe operates in a multi-body dynamical environment and must perform observation and downlink tasks under strict geometric, visibility, and communication constraints. Furthermore, mission execution is complicated by additional uncertainties and operational requirements, including stochastic cloud occlusion, solar avoidance constraints, and time-varying scientific utility of candidate targets. In addition, probes must continuously coordinate strongly coupled onboard resources under mutually exclusive observation and transmission modes. 
Under such conditions, these traditional methods show clear limitations in highly uncertain and dynamically changing deep space environments. Rule-based heuristic methods usually rely heavily on handcrafted designs, metaheuristic methods are often parameter-sensitive and prone to local optima, and multi-objective evolutionary algorithms still struggle to capture the complex mapping between uncertainty, mission states, and planning decisions. Therefore, they are inadequate for deep space autonomous mission planning that requires rapid response and online adaptation under uncertainty.

To address the above limitations of traditional optimization methods, researchers have increasingly introduced reinforcement learning (RL) into mission planning and scheduling problems \cite{su2024novel,yang2023impact}. RL has shown strong potential for modeling multiple sequential decision processes and resource-constrained operations in spacecraft planning and Earth-observing satellite scheduling \cite{zhu2025constrained,qiu2021scheduling}. For example, He et al. \cite{he2020generic} formulated agile Earth observation satellite scheduling as a generic Markov decision process (MDP) and solved it using RL, while Wang et al. \cite{wang2022deep} developed a deep reinforcement learning (DRL) method for autonomous collaborative mission planning of high- and low-orbit agile Earth-observing satellites. Wen et al. \cite{wen2023scheduling} further incorporated a hybrid actor-critic framework to jointly schedule observation and transmission tasks. Herrmann et al. \cite{herrmann2023reinforcement} investigated RL for on-board scheduling of agile Earth-observing satellites, and subsequent studies extended this line toward scalable multi-satellite operations and distributed cooperation. Building on these developments, Liang et al. \cite{liang2023fast} proposed a DRL approach for rapid satellite range rescheduling. Li et al. \cite{li2024mission} introduced an attention-based DRL method for distributed mission planning of multiple agile Earth-observing satellites. More recently, DRL has continued to evolve toward more structured and adaptive mission planners. Wang et al. \cite{wang2025deep} proposed an improved DRL framework for agile Earth observation satellite scheduling with variable image duration. Zhou et al. \cite{zhou2025deep} developed a two-phase hybrid DRL framework for multi-agile Earth observation satellite scheduling, and Li et al. \cite{li2025mission} presented a mission sequence model with DRL for dynamic multi-satellite replanning. Meanwhile, Qi et al. \cite{qi2024cooperative} formulated an MDP model integrating mission planning and proposed the PB-PPO algorithm. Yao et al.~\cite{yao2026meta} investigated a meta-reinforcement learning method for dynamic mission scheduling of Earth observation satellites. Li and Dong \cite{xuedong2025expert} further proposed an expert-guided hierarchical RL framework for collaborative mission planning in low Earth orbit (LEO) satellite clusters, improving policy interpretability and robustness under uncertainty. 
However, most existing RL-based studies still focus on LEO satellite scenarios, whereas autonomous mission planning for deep space exploration in the Saturnian system remains far less explored. Moreover, many existing methods rely on relatively simplified dynamics and uncertainty patterns and still encounter substantial difficulties in representing high-dimensional mission states, handling strongly constrained and dynamically varying feasible action sets, and optimizing long-horizon sequential decision-making processes. In particular, opportunity targets in deep space missions may become available during mission execution. These limitations restrict their applicability to autonomous Titan mission planning under complex deep space conditions.

To address the above issues, we propose a DRL method, termed physics-constrained maskable proximal policy optimization (PCM-PPO), for Titan deep space probes. Specifically, a discrete-time Markov decision process (MDP) is formulated to characterize the coupled physical constraints, resource evolution, dynamic target availability, and uncertainties in Titan mission planning. Within the PCM-PPO framework, a fixed-capacity target pool is introduced to represent predefined targets and dynamically arriving opportunity targets under a bounded action dimension. Based on this representation, the mission state is organized into a target-wise state matrix and a global mission-state vector, which enables target-level attributes and global mission conditions to be represented separately. The actor network generates action logits through a shared target encoder, a context network, and a target-scoring function, so that active targets are evaluated according to their current mission attributes and global planning context. In addition, a physics-constrained action masking mechanism is introduced to construct the admissible decision domain from Titan-specific mission states, including target-slot occupancy, geometric observability, Deep Space Network (DSN) availability, task-completion status, solar safety, cloud visibility, and resource safety. Thus, action sampling and policy updating are restricted to physically executable decisions rather than the full unconstrained action space. An integrated training strategy is further employed to facilitate policy optimization under high-dimensional mission states and strongly coupled constraints. The main contributions of this paper are summarized as follows:

(1) A Titan-specific discrete-time MDP is formulated for autonomous Titan mission planning. By incorporating a system dynamics and geometric model with $J_2$ perturbation, third-body attraction, and downlink occultation, the formulation integrates multi-body orbital dynamics, observation and downlink feasibility, onboard resource evolution, dynamic target availability, time-varying target utility, and environmental uncertainties into a unified planning framework.

(2) The PCM-PPO method is introduced by integrating target pool-based state representation, target-conditioned policy modeling, and a Titan-specific physics-constrained action mask. A fixed-capacity target pool representation is introduced to accommodate predefined targets and dynamically arriving opportunity targets under a bounded action dimension. Based on this representation, the mission state is organized into a target-wise state matrix and a global mission-state vector, and the actor generates action logits through a shared target encoder, a context network, and a target-scoring function. 

(3) A Titan-specific physics-constrained action mask is constructed from the current mission state to define the admissible decision domain. By encoding target-slot availability, observation geometry, DSN availability, stochastic visibility, task-completion status, solar safety, and resource safety constraints into a unified feasibility mapping, the mask restricts action sampling and policy updating to physically executable decisions and reduces ineffective exploration over infeasible actions.

(4) An integrated training strategy is also incorporated to support policy optimization under high-dimensional mission states and strongly coupled constraints. Comprehensive simulations are conducted under varying target loads, long-horizon evaluation scenarios, ablation settings, and uncertain mission conditions. The experimental results validate the effectiveness of PCM-PPO for autonomous mission planning of Titan deep space probes.

\begin{figure*}[!t]
  \centering
  \includegraphics[width=0.66\textwidth]{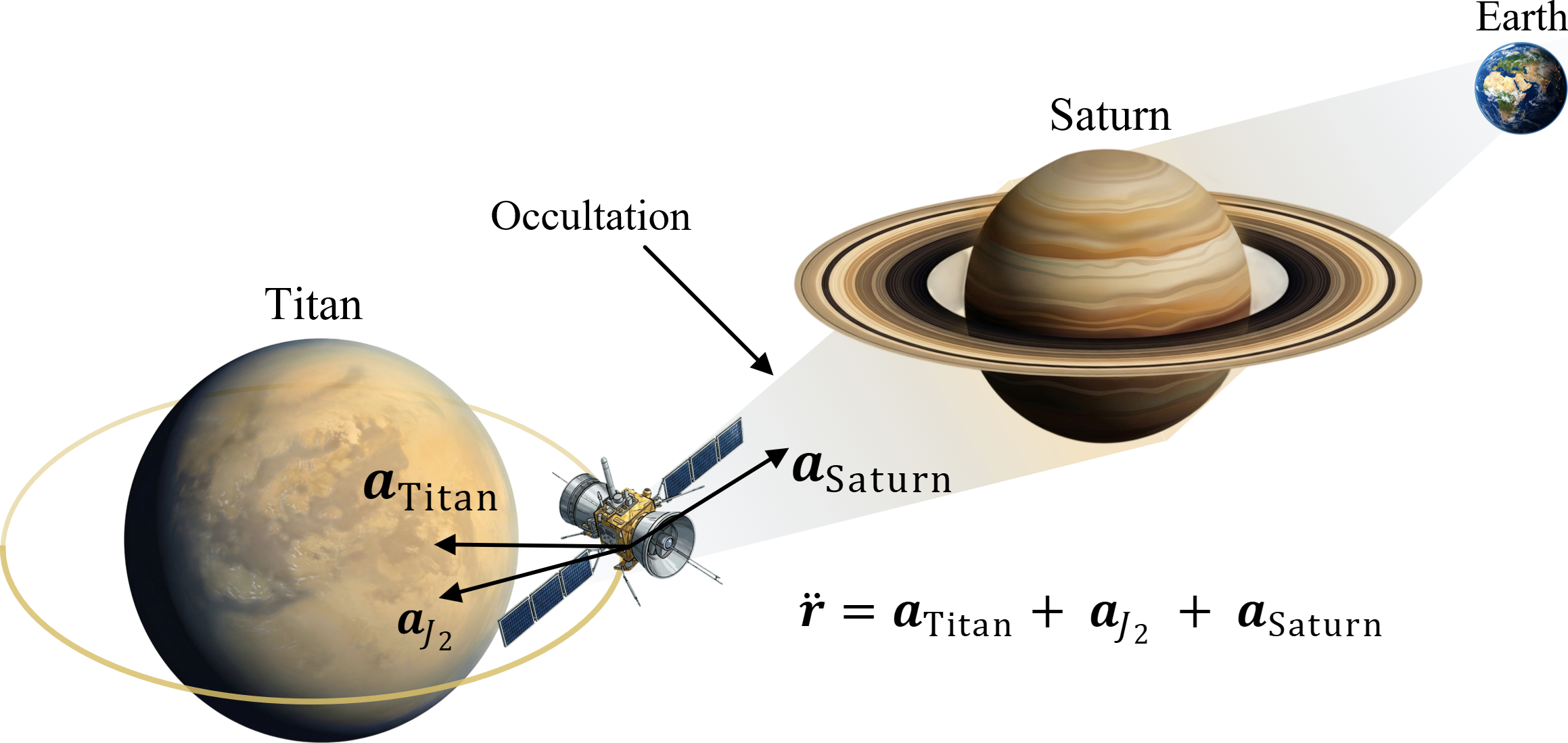}
  \caption{Multi-body dynamics and geometric constraints in the Titan mission environment.}
  \label{fig_2_1}
\end{figure*}

The remainder of this paper is organized as follows. \secref{2} formulates the Titan autonomous mission planning problem, including the system dynamics and geometric model, environmental uncertainty and resource model, and logical feasibility and operational constraints. \secref{3} introduces the PCM-PPO method, including the Markov decision process formulation, physics-constrained action masking mechanism, network structure, and integrated training process. \secref{4} presents the experimental studies, including training convergence, comparative evaluations under varying target loads, long-horizon tests, ablation and sensitivity analyses, and planning adaptability evaluations under uncertain mission conditions. Finally, \secref{5} concludes this paper and discusses future research directions.

\section{Problem formulation}
\label{2}

\subsection{System dynamics and geometric model}\label{2.1}

In Titan exploration missions, the executability of observation and downlink actions depends jointly on orbital propagation, communication line-of-sight conditions, and target observation geometry. Different from conventional LEO satellites, a Titan deep space probe evolves in a Titan-Saturn multi-body environment, where the probe motion is influenced not only by Titan’s central gravity but also by the \(J_2\) perturbation and Saturn’s third-body attraction. Meanwhile, Earth downlink opportunities may be occulted by either Titan or Saturn, and target observability is further constrained by horizon visibility, sensor field-of-view, and solar illumination conditions. Accordingly, the deep space probes' dynamics are formulated in the Titan-Centered Inertial (TCI) frame, and the geometric feasibility of observation and downlink actions is explicitly modeled, as schematically illustrated in \figref{fig_2_1}.

Let the probe state be defined as
\begin{equation}
\bm{x}(t) = \begin{bmatrix}
\bm{r}(t)\\
\bm{v}(t)
\end{bmatrix}
= [x(t), y(t), z(t), v_x(t), v_y(t), v_z(t)]^\top,
\end{equation}
where $\bm{r}(t)$ and $\bm{v}(t)$ denote the probe position and velocity vectors, respectively. The orbital dynamics are governed by the combined effects of Titan's central gravity, Titan's oblateness represented by the $J_2$ zonal harmonic, and Saturn's third-body perturbation. Since the probe flies at an altitude higher than $1000$ km, atmospheric drag at the considered orbital altitude can be ignored \cite{standley2006cassini,matson2002cassini}, and the translational dynamics are written as
\begin{equation}
\ddot{\bm{r}} = \bm{a}_{\text{Titan}} + \bm{a}_{J_2} + \bm{a}_{\text{Saturn}} .
\end{equation}

Accordingly, the central-body and $J_2$ accelerations in the Titan-centered frame can be expressed as follows:
\begin{equation}
\ddot{x} = -\frac{\mu_{\mathrm{T}} x}{r^3} \left[ 1 + \frac{3}{2} J_{2,\mathrm{T}} \left( \frac{R_{\mathrm{T}}}{r} \right)^2 \left( 1 - 5 \frac{z^2}{r^2} \right) \right],
\end{equation}

\begin{equation}
\ddot{y} = -\frac{\mu_{\mathrm{T}} y}{r^3} \left[ 1 + \frac{3}{2} J_{2,\mathrm{T}} \left( \frac{R_{\mathrm{T}}}{r} \right)^2 \left( 1 - 5 \frac{z^2}{r^2} \right) \right],
\end{equation}

\begin{equation}
\ddot{z} = -\frac{\mu_{\mathrm{T}} z}{r^3} \left[ 1 + \frac{3}{2} J_{2,\mathrm{T}} \left( \frac{R_{\mathrm{T}}}{r} \right)^2 \left( 3 - 5 \frac{z^2}{r^2} \right) \right],
\end{equation}
where $r=\|\bm{r}\|$, $\mu_{\mathrm{T}}$ is Titan's gravitational parameter, $R_{\mathrm{T}}$ is Titan's mean radius, and $J_{2,\mathrm{T}}$ is Titan's second zonal harmonic coefficient. Since long-duration missions in the Saturnian system are also influenced by Saturn's gravity, the third-body perturbation is modeled as

\begin{equation}
\bm{a}_{\text{Saturn}} =
\mu_{\mathrm{S}} \left(
\frac{\bm{r}_{\mathrm{S}} - \bm{r}}{\|\bm{r}_{\mathrm{S}} - \bm{r}\|^3}
-
\frac{\bm{r}_{\mathrm{S}}}{\|\bm{r}_{\mathrm{S}}\|^3}
\right),
\end{equation}
where $\mu_{\mathrm{S}}$ is Saturn's gravitational parameter and $\bm{r}_{\mathrm{S}}$ denotes Saturn's position relative to Titan in the TCI frame.

In addition to orbital evolution, Earth downlink feasibility must be explicitly modeled because the communication line of sight may be occulted by either Titan or Saturn. Let $\bm{r}_{\mathrm{pr}}$ denote the probe position, $\bm{r}_{\oplus}$ the Earth position, and $\bm{r}_B$ the center of an occulting body $B\in\{\mathrm{Titan},\mathrm{Saturn}\}$. The Earth line-of-sight vector and the relative vector from the probe to the occulting body are respectively defined as
\begin{equation}
\bm{u}_{\oplus}=\bm{r}_{\oplus}-\bm{r}_{\mathrm{pr}},\qquad
\bm{u}_B=\bm{r}_B-\bm{r}_{\mathrm{pr}}.
\end{equation}
The angular separation between the Earth direction and occulting body $B$ is
\begin{equation}
\theta_B=
\arccos
\left(
\frac{\bm{u}_{\oplus}\cdot\bm{u}_B}
{\|\bm{u}_{\oplus}\|\,\|\bm{u}_B\|}
\right),
\end{equation}
and the apparent angular radius of the occulting body is
\begin{equation}
\psi_B=
\arcsin\left(\frac{R_B}{\|\bm{u}_B\|}\right),
\end{equation}
where $R_B$ is the mean radius of occulting body $B$. The Earth downlink is regarded as occulted by body $B$ when the Earth direction falls within the apparent disk of the body and Earth is farther away than the occulting body, namely,
\begin{equation}
\chi_B=
\begin{cases}
0, & \theta_B<\psi_B \;\text{and}\; \|\bm{u}_{\oplus}\|>\|\bm{u}_B\|,\\
1, & \text{otherwise}.
\end{cases}
\end{equation}
Therefore, the final downlink feasibility is determined by the conjunction of Titan occlusion, Saturn occlusion, and the Deep Space Network elevation requirement:
\begin{equation}
C_{\mathrm{dl}}=
\chi_{\mathrm{Titan}}
\wedge
\chi_{\mathrm{Saturn}}
\wedge
(\epsilon_{\mathrm{DSN}}\ge \epsilon_{\min}),
\end{equation}
where $\epsilon_{\mathrm{DSN}}$ is the DSN elevation angle and $\epsilon_{\min}$ is the minimum allowable elevation threshold. In the full scheduling environment, the executable downlink action is further gated by an administrative DSN contact window, which is modeled separately in the mission environment.

Scientific targets on Titan are treated as individual observation units rather than extended regional coverage objects. For a target located at latitude $\phi$ and longitude $\lambda$, its inertial position in the TCI frame is expressed as
\begin{equation}
\bm{r}_{\mathrm{tar}}(t)
=
\boldsymbol{R}_{\mathrm{rot}}(t)
\begin{bmatrix}
R_{\mathrm{T}}\cos\phi\cos\lambda\\
R_{\mathrm{T}}\cos\phi\sin\lambda\\
R_{\mathrm{T}}\sin\phi
\end{bmatrix},
\end{equation}
\begin{equation}
\boldsymbol{R}_{\mathrm{rot}}(t) = 
\begin{bmatrix}
\cos(\omega_{\mathrm{T}} t + \theta_0) & -\sin(\omega_{\mathrm{T}} t + \theta_0) & 0 \\
\sin(\omega_{\mathrm{T}} t + \theta_0) & \cos(\omega_{\mathrm{T}} t + \theta_0) & 0 \\
0 & 0 & 1
\end{bmatrix},
\end{equation}
where $\boldsymbol{R}_{\mathrm{rot}}(t)$ is Titan's rotational transformation matrix, $\omega_{\mathrm{T}}$ is Titan's rotation rate, and $\theta_0$ is the initial phase angle. Based on this representation, a target is considered observable only when the probe-target geometry, sensor pointing capability, and illumination condition are simultaneously satisfied. The geometric visibility requirement is enforced by
\begin{equation}
\frac{(\bm{r}_{\mathrm{pr}}-\bm{r}_{\mathrm{tar}})\cdot \bm{r}_{\mathrm{tar}}}
{\|\bm{r}_{\mathrm{pr}}-\bm{r}_{\mathrm{tar}}\|\|\bm{r}_{\mathrm{tar}}\|}>0,
\end{equation}
which ensures that the target lies on the visible hemisphere as seen from the probe. Moreover, the target direction must fall inside the payload field of view. Let $\theta_{\mathrm{lr}}$ and $\theta_{\mathrm{fb}}$ denote the cross-track and along-track viewing angles, respectively. The corresponding pointing constraints are written as
\begin{equation}
|\theta_{\mathrm{lr}}|\le \alpha_{\mathrm{lr}},\qquad
|\theta_{\mathrm{fb}}|\le \alpha_{\mathrm{fb}},
\end{equation}
where $\alpha_{\mathrm{lr}}$ and $\alpha_{\mathrm{fb}}$ are the allowable field-of-view limits. Since optical observation also depends on solar illumination, the solar elevation angle at the target location is defined as
\begin{equation}
\gamma_{\odot}
=
\frac{\pi}{2}
-
\arccos
\left(
\frac{
\bm{r}_{\mathrm{tar}}\cdot(\bm{r}_{\odot}-\bm{r}_{\mathrm{tar}})
}{
\|\bm{r}_{\mathrm{tar}}\|\,\|\bm{r}_{\odot}-\bm{r}_{\mathrm{tar}}\|
}
\right),
\end{equation}
where $\bm{r}_{\odot}$ denotes the Sun position vector in the TCI frame. The illumination requirement is then given by
\begin{equation}
\gamma_{\odot}>\theta_{\odot,\min},
\end{equation}
where $\theta_{\odot,\min}$ is the minimum allowable solar elevation angle. An observation opportunity is feasible if horizon visibility, payload pointing, and illumination conditions are all satisfied simultaneously.

For off-nadir observations, the lateral pointing side is determined from the relative orientation among the probe position, velocity, and target direction. Let
\begin{equation}
\bm{u}_{\mathrm{tar}}=\bm{r}_{\mathrm{tar}}-\bm{r}_{\mathrm{pr}},
\end{equation}
and define the roll-side indicator as
\begin{equation}
\sigma_{\mathrm{roll}}=
\begin{cases}
+1, & (\bm{r}_{\mathrm{pr}}\times \bm{v}_{\mathrm{pr}})\cdot \bm{u}_{\mathrm{tar}}>0,\\
-1, & \mathrm{otherwise}.
\end{cases}
\end{equation}
where $\bm{r}_{\mathrm{pr}}$ and $\bm{v}_{\mathrm{pr}}$ denote the probe position and velocity vectors, respectively. The indicator $\sigma_{\mathrm{roll}}$ is used to identify the feasible lateral pointing direction for off-nadir observation.

\subsection{Environmental uncertainty and resource model}\label{2.2}
We further model the environmental uncertainty and resource dynamics in Titan mission planning. Specifically, solar safety, stochastic cloud occlusion, time-varying task utility, and the coupled evolution of onboard energy, data storage, and communication opportunity are modeled in discrete time.

For optical payloads, we impose an additional sun-avoidance requirement to protect the onboard optical payload from excessive solar exposure. Let
\begin{equation}
\bm{d}_i(t)
=
\bm{r}_{\mathrm{tar},i}(t)
-
\bm{r}_{\mathrm{pr}}(t)
\end{equation}
denote the line-of-sight vector from the probe to target $\tau_i$, and let $\bm{u}_{\odot}(t)$ denote the Sun direction vector in the same reference frame. Accordingly, the corresponding safety indicator is defined as
\begin{equation}
I_{\mathrm{safe},i}(t)
=
\begin{cases}
1, &
\dfrac{
\bm{d}_i(t)
\cdot
\bm{u}_{\odot}(t)
}{
\|\bm{d}_i(t)\|\,
\|\bm{u}_{\odot}(t)\|
}
\le
\cos \phi_{\mathrm{safe}},
\\[8pt]
0, & \text{otherwise},
\end{cases}
\end{equation}
where $I_{\mathrm{safe},i}(t)=1$ means that target $\tau_i$ satisfies the solar safety requirement at time $t$, and $\phi_{\mathrm{safe}}$ is the minimum allowable angular separation. For notational clarity, $i$ denotes a target-pool slot, $k$ denotes a discrete decision step, and $t_k$ denotes the physical time at decision step $k$. The occupancy state of slot $i$ at continuous time $t$ is denoted by $q_i(t)$, and its value at decision step $k$ is denoted by $q_{i,k}$.

Besides solar safety, Titan observations are also affected by stochastic cloud occlusion. For each candidate target $\tau_i$, we introduce a cloud-free probability state $\varpi_i(t)\in[0,1]$ and a binary visibility realization $\xi_i(t)\in\{0,1\}$. The probability state evolves as
\begin{equation}
\mu_{\mathrm{eff}}
=
\mu_{\mathrm{c}}^{\Delta t/\Delta t_{\mathrm{c}}},
\end{equation}
\begin{equation}
\varpi_i(t+\Delta t)
=
\mathrm{clip}
\left(
\mu_{\mathrm{eff}}\varpi_i(t)
+
(1-\mu_{\mathrm{eff}})\bar{\varpi}_i
+
\epsilon_i(t),\,0,\,1
\right),
\end{equation}
where $\mu_{\mathrm{c}}\in(0,1)$ is the cloud persistence coefficient,
$\Delta t$ is the planning step, $\Delta t_{\mathrm{c}}$ is the cloud-update
reference interval, $\bar{\varpi}_i$ is the climatological cloud-free
probability of target $\tau_i$, and
\begin{equation}
\epsilon_i(t)
\sim
\mathcal{N}
\left(
0,\,
\sigma_{\mathrm{c}}^2
\frac{\Delta t}{\Delta t_{\mathrm{c}}}
\right)
\end{equation}
is the step-scaled stochastic perturbation. The actual cloud visibility realization is sampled as
\begin{equation}
\xi_i(t)
\sim
\mathrm{Bernoulli}
\!\big(
\varpi_i(t)
\big),
\end{equation}
where $\xi_i(t)=1$ indicates that target $\tau_i$ is observable.

The scientific utility of mission targets is also time-varying, especially for opportunity targets that may arrive during mission execution and rapidly lose importance if not observed in time. To model dynamically arriving targets while keeping the decision dimension bounded, a fixed-capacity target pool is introduced as
\begin{equation}
\mathcal{T}_{\mathrm{pool}}
=
\{
\tau_1,\tau_2,\ldots,\tau_{K_{\max}}
\},
\end{equation}
where $K_{\max}=N_{\mathrm{base}}+N_{\mathrm{opp}}^{\max}$ is the maximum target-pool capacity, $N_{\mathrm{base}}$ is the number of predefined mapping targets, and $N_{\mathrm{opp}}^{\max}$ is the maximum number of opportunity-target slots. For each target slot $\tau_i$, an occupancy indicator $q_i(t)\in\{0,1\}$ is defined, where $q_i(t)=1$ indicates that the slot is occupied by an active target and $q_i(t)=0$ indicates that the slot is empty. Predefined mapping targets are initialized at the beginning of the mission, whereas opportunity targets are inserted into empty slots only after their arrival. Therefore, an opportunity target does not participate in state representation, action selection, or policy updating before it is inserted into the target pool.

The instantaneous utility of target $\tau_i$ is defined as
\begin{equation}
U_i(t)
=
\begin{cases}
0, & q_i(t)=0,
\\[4pt]
U_i^{\mathrm{init}}
\exp
\!\left[
-\lambda_i
\left(
t-t_i^{\mathrm{arr}}
\right)
\right],
& q_i(t)=1,
\end{cases}
\end{equation}
where $t_i^{\mathrm{arr}}$ is the arrival time of target $\tau_i$, $U_i^{\mathrm{init}}$ is its initial utility, and $\lambda_i$ is the corresponding decay rate. For predefined mapping targets, $t_i^{\mathrm{arr}}$ is set to the initial mission time and $\lambda_i\approx 0$. For transient opportunity targets, $t_i^{\mathrm{arr}}$ denotes the time when the target is inserted into the dynamic target pool, and larger decay rates are assigned to reflect stronger temporal urgency.

Based on the above environmental factors, the effective executable task set at time $t$ is defined as
\begin{equation}
\begin{aligned}
\mathcal{T}_{\mathrm{feas}}(t)
=
\bigl\{
\tau_i \in \mathcal{T}_{\mathrm{pool}}
\mid\;&
q_i(t)=1
\wedge
I_{\mathrm{geom},i}(t)
\wedge
I_{\mathrm{safe},i}(t)
\wedge
\xi_i(t)
\\
&\wedge
[U_i(t)\ge U_{\min}]
\bigr\}.
\end{aligned}
\end{equation}
where $q_i(t)=1$ ensures that only active targets in the target pool can enter the executable task set, $I_{\mathrm{geom},i}(t)$ is the geometric feasibility indicator established in \secref{2.1}, and $U_{\min}$ is the minimum utility threshold required to justify resource consumption. This definition explicitly separates target-pool occupancy, geometric visibility, solar safety, weather realization, and task-value filtering.

Let $a_t$ denote the action executed at time $t$. Since the probe is powered by a radioisotope thermoelectric generator (RTG), the onboard energy state $E(t)$ evolves as
\begin{equation}
E(t+\Delta t)
=
E(t)
+
\frac{
\left(
P_{\mathrm{RTG}}
-
P_{\mathrm{base}}
\right)
\Delta t
}{3600}
-
e(a_t),
\end{equation}
where $P_{\mathrm{RTG}}$ and $P_{\mathrm{base}}$ denote the RTG output power and baseline system power, respectively, in W, and $e(a_t)$ denotes the action-dependent energy cost in Wh. Here, $\Delta t$ is expressed in s, and the factor $1/3600$ converts the net power contribution over one decision step from $\mathrm{W\cdot s}$ to $\mathrm{Wh}$.

Let $S(t)$ denote the onboard data storage state. The data volume generated at time $t$ is denoted by $d(a_t)$. The actual downlinked data volume is
\begin{equation}
\delta_{\mathrm{down}}(t)
=
\begin{cases}
\min\left(S(t), R_{\mathrm{down}}(t)\Delta t\right),
& a_t=a^{\mathrm{dl}},\ I_{\mathrm{comm}}(t)=1,\\
0, & \mathrm{otherwise},
\end{cases}
\end{equation}
and the storage state evolves as
\begin{equation}
S(t+\Delta t)=S(t)+d(a_t)-\delta_{\mathrm{down}}(t),
\end{equation}
where $S_{\max}$ is the storage capacity limit.
The instantaneous downlink rate is modeled as
\begin{equation}
R_{\mathrm{down}}(t)=
R_{\mathrm{ref}}
\left(
\frac{D_{\oplus,\mathrm{ref}}}{D_{\mathrm{Earth}}(t)}
\right)^2,
\end{equation}
where $R_{\mathrm{ref}}$ is the reference downlink rate at distance $D_{\oplus,\mathrm{ref}}$, and $D_{\mathrm{Earth}}(t)$ is the Earth--probe distance at time $t$. Let $W_{\mathrm{DSN}}$ denote the allocated DSN contact window set. The communication availability indicator is defined as
\begin{equation}
I_{\mathrm{comm}}(t)=
C_{\mathrm{dl}}(t)\,\mathbf{I}\!\big(t\in W_{\mathrm{DSN}}\big),
\end{equation}
where $C_{\mathrm{dl}}(t)$ is the physical downlink feasibility established in \secref{2.1} and $\mathbf{I}(\cdot)$ is the indicator function.

\subsection{Logical feasibility and operational constraints}\label{2.3}

The mission horizon is discretized into $M$ decision steps with step size $\Delta t$. At each step, the probe executes one primary operational mode among target observation, data downlink, and idle standby. Since the target set is represented by the bounded dynamic target pool introduced in \secref{2.2}, observation decisions are defined over the maximum target-pool capacity $K_{\max}$. Accordingly, we introduce the binary execution indicators $y_i^{\mathrm{obs}}(t)\in\{0,1\}$, $y^{\mathrm{dl}}(t)\in\{0,1\}$, and $y^{\mathrm{idle}}(t)\in\{0,1\}$, where $y_i^{\mathrm{obs}}(t)=1$ indicates that the target currently stored in slot $\tau_i$ is selected for observation, $y^{\mathrm{dl}}(t)=1$ indicates that a downlink task is executed, and $y^{\mathrm{idle}}(t)=1$ indicates that the probe remains in idle standby mode.

The mission execution rules are summarized as follows.

(1) Only one primary operation can be executed at each decision step:
\begin{equation}
\sum_{i=1}^{K_{\max}} y_i^{\mathrm{obs}}(t)
+
y^{\mathrm{dl}}(t)
+
y^{\mathrm{idle}}(t)
=
1.
\end{equation}

(2) A target observation can be executed only when the corresponding target slot is occupied, the target belongs to the executable task set, and the target has not been completed before. Let $c_i(t)\in\{0,1\}$ denote the completion flag of target slot $\tau_i$, where $c_i(t)=1$ means that the active target in this slot has already been completed. Then
\begin{equation}
\begin{aligned}
y_i^{\mathrm{obs}}(t)
&\le
q_i(t),
\\
y_i^{\mathrm{obs}}(t)
&\le
\mathbf{I}
\!\left(
\tau_i\in\mathcal{T}_{\mathrm{feas}}(t)
\right),
\\
y_i^{\mathrm{obs}}(t)
&\le
1-c_i(t),
\qquad
i=1,\ldots,K_{\max}.
\end{aligned}
\end{equation}
The first inequality ensures that an empty target slot cannot be selected for observation. The second inequality enforces the physical and environmental feasibility conditions encoded in $\mathcal{T}_{\mathrm{feas}}(t)$ in Eq.~(28), and the third inequality prevents repeated observation of completed targets.

The completion indicator is expressed as
\begin{equation}
c_i(t+\Delta t)
=
\min
\left(
c_i(t)
+
y_i^{\mathrm{obs}}(t),
1
\right),
\qquad
i=1,\ldots,K_{\max}.
\end{equation}
For an empty target slot with $q_i(t)=0$, the corresponding observation action is inadmissible and its completion flag does not affect the mission state.

(3) A downlink task can be executed only when the communication link is available and there is valid data stored onboard:
\begin{equation}
y^{\mathrm{dl}}(t)
\le
I_{\mathrm{comm}}(t),
\qquad
y^{\mathrm{dl}}(t)
\le
\mathbf{I}
\!\left(
S(t)>0
\right).
\end{equation}

(4) Let $\hat{E}(t+\Delta t\mid a_t)$ and $\hat{S}(t+\Delta t\mid a_t)$ denote the one-step predicted energy and storage states under action $a_t$. The resource-safe action set is defined as
\begin{equation}
a_t
\in
\mathcal{A}_{\mathrm{safe}}(t)
\Longleftrightarrow
\begin{cases}
E_{\min}
+
\Delta E_{\mathrm{safe}}
\le
\hat{E}(t+\Delta t\mid a_t)
\le
E_{\max},
\\[3pt]
0
\le
\hat{S}(t+\Delta t\mid a_t)
\le
S_{\max}
-
\Delta S_{\mathrm{safe}},
\end{cases}
\end{equation}
where $E_{\min}$ denotes the minimum allowable onboard energy, and $\Delta E_{\mathrm{safe}}$ and $\Delta S_{\mathrm{safe}}$ denote the prescribed safety margins for energy and storage, respectively.

\section{Physics-constrained maskable proximal policy optimization method}\label{3}

\subsection{Markov decision process formulation}\label{3.1}

\begin{figure}[!t]
  \centering
  \includegraphics[width=7.0cm]{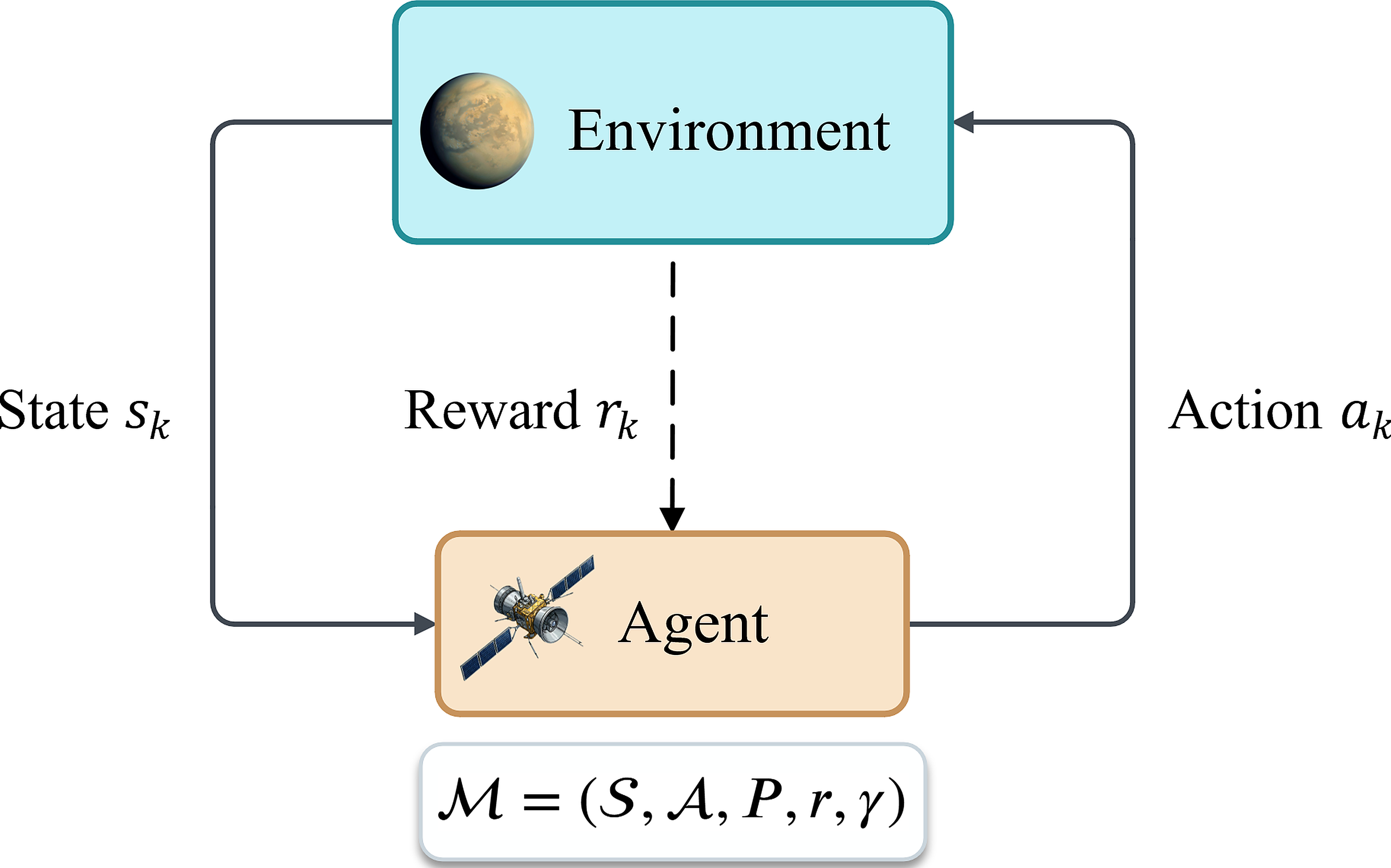}
  \caption{Markov decision process formulation for Titan mission planning.}
  \label{fig_3.1}
\end{figure}

RL provides a natural framework for sequential decision-making problems in which an agent repeatedly interacts with an environment and improves its policy through trial-and-error experience \cite{nguyen2020deep,cao2023reinforcement}. RL methods can generally be divided into two categories: value-based methods and policy-based methods. Value-based methods learn action values for decision-making, whereas policy-based methods directly optimize the policy and are more suitable for complex sequential decision problems with high-dimensional state spaces \cite{matsuo2022deep}.
As illustrated in \figref{fig_3.1}, the agent observes the current state of the environment, selects an action according to its policy, and then receives the corresponding reward and next state through environmental feedback. For autonomous Titan mission planning, the probe must repeatedly select observation, downlink, or standby actions under time-varying mission opportunities, environmental uncertainty, and coupled onboard resource constraints.

The autonomous mission planning process of the Titan probe is formulated as a Markov decision process (MDP). We model the probe as an agent, and each decision step corresponds to one interaction between the onboard planner and the mission environment. The MDP is written as
\begin{equation}
\mathcal{M}=(\mathcal{S},\mathcal{A},P,r,\gamma),
\end{equation}
where $\mathcal{S}$ denotes the state space, $\mathcal{A}$ the action space, $P$ the state transition kernel, $r$ the immediate reward function, and $\gamma\in(0,1]$ the discount factor.

The state space is denoted by $\mathcal{S}$. Since the target set is represented by the bounded dynamic target pool introduced in \secref{2.2}, the mission state is organized into target-level features and global mission-state features. At decision step $k$, the state observed by the agent is represented as
\begin{equation}
s_k=
\left(
\bm{X}_k,\bm{g}_k
\right),
\end{equation}
where $\bm{X}_k$ is the target-feature matrix and $\bm{g}_k$ is the global mission-state vector.

The target-feature matrix is defined as
\begin{equation}
\bm{X}_k
=
\left[
\bm{x}_{1,k},
\bm{x}_{2,k},
\ldots,
\bm{x}_{K_{\max},k}
\right]^{\top}
\in \mathbf{R}^{K_{\max}\times d_x},
\end{equation}
where $\bm{x}_{i,k}$ denotes the target-level feature vector of slot $i$ at decision step $k$, and $d_x$ is the dimension of each target-level feature vector. Specifically, $\bm{x}_{i,k}$ is given by
\begin{equation}
\begin{aligned}
\bm{x}_{i,k}
=
\big[
&q_{i,k},\ \tilde{U}_{i,k},\ \varpi_{i,k},\
I_{\mathrm{geom},i,k},\ I_{\mathrm{safe},i,k},\\
&\xi_{i,k},\ c_{i,k},\ \eta_{i,k}
\big]^{\top},\\
&i=1,\ldots,K_{\max}.
\end{aligned}
\end{equation}
where $q_{i,k}=q_i(t_k)$ denotes the occupancy indicator of target slot $i$, $\tilde{U}_{i,k}$ is the normalized utility of the target currently stored in this slot, $\varpi_{i,k}$ is the cloud-free probability, $I_{\mathrm{geom},i,k}$ and $I_{\mathrm{safe},i,k}$ denote the geometric-feasibility and solar-safety indicators, respectively, $\xi_{i,k}$ denotes the stochastic visibility realization, $c_{i,k}=c_i(t_k)$ denotes the task-completion indicator, and $\eta_{i,k}$ denotes the normalized maneuver-demand index. For an empty target slot with $q_{i,k}=0$, the corresponding target-level features are set to zero and the slot is excluded from feasible action selection by the physics-constrained action mask.

The global mission-state vector is defined as
\begin{equation}
\bm{g}_k
=
\left[
\bar{t}_k,
\bar{E}_k,
\bar{S}_k,
\bar{R}_k^{\mathrm{down}},
\bar{D}_k^{\mathrm{Earth}},
C_{\mathrm{dl}}(t_k),
I_k^{\mathrm{DSN}}
\right]^{\top}
\in \mathbf{R}^{d_g},
\end{equation}
where $\bar{t}_k$ denotes the normalized mission progress at decision step $k$, $\bar{E}_k$ and $\bar{S}_k$ denote the normalized onboard energy and storage states, respectively, $\bar{R}_k^{\mathrm{down}}$ denotes the normalized downlink rate, $\bar{D}_k^{\mathrm{Earth}}$ denotes the normalized Earth-probe distance, $C_{\mathrm{dl}}(t_k)$ denotes the physical downlink-feasibility indicator, and $I_k^{\mathrm{DSN}}=\mathbf{I}(t_k\in W_{\mathrm{DSN}})$ indicates whether the current step lies within the allocated DSN contact window. Accordingly, the state space satisfies
\begin{equation}
s_k\in\mathcal{S}
\subseteq
\mathbf{R}^{K_{\max}\times d_x}
\times
\mathbf{R}^{d_g}.
\end{equation}

The action space is discrete and defined over the bounded target pool as
\begin{equation}
\mathcal{A}
=
\left\{
a_1^{\mathrm{obs}},
\ldots,
a_{K_{\max}}^{\mathrm{obs}},
a^{\mathrm{dl}},
a^{\mathrm{idle}}
\right\},
\end{equation}
where $a_i^{\mathrm{obs}}$ denotes selecting the target currently stored in slot $i$ for observation, $a^{\mathrm{dl}}$ denotes data downlink, $a^{\mathrm{idle}}$ denotes idle standby, and the action selected by the agent is denoted by $a_k\in\mathcal{A}$ at decision step $k$. If $q_{i,k}=0$, the corresponding observation action $a_i^{\mathrm{obs}}$ is regarded as inadmissible and is removed by the physics-constrained action mask. For maneuver-required targets, the corresponding observation action is still treated as one discrete action at the planning level, while its internal maneuver-observe-recovery process is absorbed into the environment transition.

Given the current state $s_k$ and selected action $a_k$, the next state is generated according to
\begin{equation}
s_{k+1}\sim P(\cdot\mid s_k,a_k),
\end{equation}
where the transition process jointly reflects orbital and geometric evolution, stochastic cloud variation, target utility updates, energy and storage, communication opportunity evolution, and mission completion updates.

The reward is defined as
\begin{equation}
r_k
=
r_k^{\mathrm{obs}}
+
r_k^{\mathrm{dl}}
-
r_k^{\mathrm{res}}
-
r_k^{\mathrm{idle}},
\end{equation}
where $r_k^{\mathrm{obs}}$ denotes the observation-utility reward, $r_k^{\mathrm{dl}}$ denotes the downlink reward, $r_k^{\mathrm{res}}$ denotes the resource-consumption penalty, and $r_k^{\mathrm{idle}}$ denotes the idle penalty.

For target-observation actions, the observation reward is defined over the bounded target pool as
\begin{equation}
r_k^{\mathrm{obs}}
=
\lambda_{\mathrm{obs}}
\sum_{i=1}^{K_{\max}}
\mathbf{I}
\!\left(
a_k=a_i^{\mathrm{obs}}
\right)
q_{i,k}
\left(
1-c_{i,k}
\right)
\tilde{U}_{i,k},
\end{equation}
where $\lambda_{\mathrm{obs}}$ is the observation-reward coefficient, $\mathbf{I}(\cdot)$ is the indicator function, $q_{i,k}$ is the occupancy indicator of target slot $i$, $c_{i,k}$ is the task-completion indicator, and $\tilde{U}_{i,k}$ denotes the normalized utility of the target currently stored in slot $i$ at decision step $k$. The term $q_{i,k}$ ensures that empty target slots do not contribute to the observation reward.

For downlink actions, the reward is defined according to the actually transmitted data volume,
\begin{equation}
r_k^{\mathrm{dl}}
=
\lambda_{\mathrm{dl}}
\mathbf{I}\!\left(a_k=a^{\mathrm{dl}}\right)
\frac{\delta_{\mathrm{down}}(t_k)}{S_{\max}},
\end{equation}
where $\lambda_{\mathrm{dl}}$ is the downlink-reward coefficient, $\delta_{\mathrm{down}}(t_k)$ is the actual downlinked data volume at time $t_k$, and $S_{\max}$ is the onboard storage capacity.

To discourage resource-heavy decisions, the resource-consumption penalty is defined as
\begin{equation}
r_k^{\mathrm{res}}
=
\lambda_E
\frac{e(a_k)}{E_{\max}}
+
\lambda_S
\frac{\max(S(t_{k+1})-S(t_k),0)}{S_{\max}},
\end{equation}
where $\lambda_E$ and $\lambda_S$ are the weighting coefficients for energy and storage penalties, respectively, $e(a_k)$ is the energy cost of action $a_k$, and $E_{\max}$ denotes the maximum onboard energy.

To prevent overly conservative behavior, a small idle penalty is introduced as
\begin{equation}
r_k^{\mathrm{idle}}
=
\lambda_{\mathrm{idle}}
\mathbf{I}\!\left(a_k=a^{\mathrm{idle}}\right),
\end{equation}
where $\lambda_{\mathrm{idle}}$ is a small positive constant.

Accordingly, the reward jointly reflects scientific return from active targets, effective data return, and long-horizon resource sustainability under the bounded dynamic target-pool representation.

The learning objective is to obtain an optimal policy $\pi^*$ that maximizes the expected cumulative discounted reward over the mission horizon, namely,
\begin{equation}
\pi^*=\arg\max_{\pi}\mathbb{E}_{\pi}\left[\sum_{k=0}^{M-1}\gamma^k r_k\right].
\end{equation}
where $\gamma\in(0,1]$ is the discount factor, which determines the relative importance of future rewards, and $k$ denotes the decision-step index. A larger $\gamma$ places more emphasis on long-horizon returns, whereas a smaller $\gamma$ gives greater weight to immediate rewards. $\pi$ denotes the policy, and $\mathbb{E}_{\pi}[\cdot]$ denotes the expectation over trajectories generated by following policy $\pi$. 

\subsection{Physics-constrained maskable proximal policy optimization algorithm}\label{3.2}
We develop a physics-constrained maskable proximal policy optimization algorithm for the Titan autonomous mission planning problem. The proposed method retains the actor-critic (AC) framework \cite{wei2021deep} and clipped policy-update mechanism of PPO \cite{yuan2023integrated}, while explicitly embedding physical and logical feasibility constraints into the action-selection process through a dynamic action mask.

The policy network and the value network are denoted by $\pi_{\theta}(a_k\mid s_k)$ and $V_{\phi}(s_k)$, respectively,
where $\theta$ and $\phi$ denote the parameters of the actor and critic, respectively. For policy evaluation and advantage estimation, the temporal-difference residual is written as
\begin{equation}
\delta_k
=
r_k+\gamma(1-d_k)V_{\phi}(s_{k+1})-V_{\phi}(s_k),
\end{equation}
where $d_k\in\{0,1\}$ indicates whether the episode terminates after step $k$. Based on the generalized advantage estimation (GAE) scheme, the advantage estimator is computed as
\begin{equation}
\hat{A}_k
=
\sum_{\ell=0}^{T_{\mathrm{ep}}-k-1}
(\gamma\lambda)^{\ell}\delta_{k+\ell},
\end{equation}
where $\gamma\in(0,1]$ is the discount factor, $\lambda\in[0,1]$ is the GAE parameter, and $T_{\mathrm{ep}}$ denotes the episode length. The term $\delta_{k+\ell}$ denotes the temporal-difference residual at step $k+\ell$, and $\hat{A}_k$ is the resulting advantage estimate at step $k$.

We then introduce a physics-constrained action mask to encode the admissible action set under the current mission state. The mask follows the general principle of state-dependent invalid-action masking, while its validity evaluation is instantiated as a Titan-specific physical feasibility mapping that couples the bounded target-pool state with orbital, environmental, communication, and resource constraints:
\begin{equation}
\bm{m}_k
=
\mathcal{M}_{\mathrm{phys}}
\left(s_k;\Theta_{\mathrm{Titan}}\right)
\in
\{0,1\}^{K_{\max}+2},
\end{equation}
where $\Theta_{\mathrm{Titan}}$ denotes the set of Titan mission parameters and physical feasibility rules, including target-slot occupancy, orbital geometry, downlink occultation, DSN availability, cloud visibility, solar safety, task-completion status, and onboard resource limits. This formulation acts as a hard, step-wise action-set restriction based on the current mission state and resource predictions, rather than an expected-cost or long-horizon constrained/safe RL formulation. Accordingly, the contribution of PCM-PPO lies in the problem-specific integration of this feasibility evaluator with the bounded dynamic target pool and the target-conditioned policy, rather than in state-dependent masking alone. The mask vector is written as
\begin{equation}
\bm{m}_k
=
\left[
m_{1,k}^{\mathrm{obs}},
\ldots,
m_{K_{\max},k}^{\mathrm{obs}},
m_{K_{\max}+1,k}^{\mathrm{dl}},
m_{K_{\max}+2,k}^{\mathrm{idle}}
\right]^{\top}.
\end{equation}

For target-observation actions, the mask of target slot $i$ is defined as
\begin{equation}
\begin{aligned}
m_{i,k}^{\mathrm{obs}}
={}&q_{i,k}\,
\mathbf{I}\!\left(\tau_i\in\mathcal{T}_{\mathrm{feas}}(t_k)\right)
\mathbf{I}\!\left(c_{i,k}=0\right)\\
&\times
\mathbf{I}\!\left(a_i^{\mathrm{obs}}\in
\mathcal{A}_{\mathrm{safe}}(t_k)\right),
\quad i=1,\ldots,K_{\max}.
\end{aligned}
\end{equation}
The occupancy term $q_{i,k}$ ensures that an empty target slot cannot be selected for observation. The other terms enforce the executable task condition, task-completion constraint, and resource-safety requirement, respectively.

For downlink and idle actions, the corresponding mask elements are defined as
\begin{equation}
\begin{aligned}
m_{K_{\max}+1,k}^{\mathrm{dl}}
={}&I_{\mathrm{comm}}(t_k)\,
\mathbf{I}\!\left(S(t_k)>0\right)
\mathbf{I}\!\left(a^{\mathrm{dl}}\in
\mathcal{A}_{\mathrm{safe}}(t_k)\right),\\
m_{K_{\max}+2,k}^{\mathrm{idle}}
={}&\mathbf{I}\!\left(a^{\mathrm{idle}}\in
\mathcal{A}_{\mathrm{safe}}(t_k)\right).
\end{aligned}
\end{equation}
The admissible action set is therefore written as
\begin{equation}
\mathcal{A}_{\mathrm{adm}}(s_k)
=
\left\{
a_j\in\mathcal{A}
\mid
m_{j,k}=1
\right\}.
\end{equation}

With the admissible action domain determined, the actor generates logits through a target-conditioned representation. Each target slot is first encoded by a shared target encoder,
\begin{equation}
\bm{h}_{i,k}
=
g_{\psi}(\bm{x}_{i,k}),
\quad i=1,\ldots,K_{\max},
\end{equation}
where $g_{\psi}(\cdot)$ is shared by all target slots. A set-level mission context is then obtained by aggregating the active target embeddings and combining them with the global mission-state vector:
\begin{equation}
\bm{c}_k
=
\rho
\left(
\sum_{i=1}^{K_{\max}} q_{i,k}\bm{h}_{i,k},
\bm{g}_k
\right),
\end{equation}
where $\rho(\cdot)$ denotes the context network. The logit of each target-observation action is generated by a shared target-scoring function:
\begin{equation}
z_{i,k}
=
f_{\theta}
\left(
\bm{h}_{i,k},
\bm{c}_k
\right),
\quad i=1,\ldots,K_{\max}.
\end{equation}
The logits for downlink and idle actions are computed from the same mission context:
\begin{equation}
z_{K_{\max}+1,k}
=
f_{\theta}^{\mathrm{dl}}(\bm{c}_k),
\qquad
z_{K_{\max}+2,k}
=
f_{\theta}^{\mathrm{idle}}(\bm{c}_k).
\end{equation}
Thus, the full logit vector is expressed as
\begin{equation}
\bm{z}_k
=
\left[
z_{1,k},
\ldots,
z_{K_{\max},k},
z_{K_{\max}+1,k},
z_{K_{\max}+2,k}
\right]^{\top}.
\end{equation}

Before the softmax operation, the logits are filtered by the physics-constrained mask:
\begin{equation}
\bar{z}_{j,k}
=
\begin{cases}
z_{j,k}, & m_{j,k}=1,\\
-\infty, & m_{j,k}=0,
\end{cases}
\quad j=1,\ldots,K_{\max}+2.
\end{equation}
The resulting masked policy distribution is defined as
\begin{equation}
\tilde{\pi}_{\theta}
(a_k=j\mid s_k,\bm{m}_k)
=
\frac{\exp(\bar{z}_{j,k})}
{\sum_{\ell=1}^{K_{\max}+2}
\exp(\bar{z}_{\ell,k})}.
\end{equation}
Since inactive target slots and physically infeasible actions are suppressed at the logit level, the resulting policy assigns probability only within the admissible action domain derived from the current Titan mission state. Consequently, policy optimization is performed over the physically executable decision domain rather than over the full unconstrained action space.

\begin{algorithm}[t]
\caption{Physics-constrained action mask construction}
\label{alg:mask_full}
\begin{algorithmic}[1]
\State \textbf{Input.} Current state $s_k=(\bm{X}_k,\bm{g}_k)$; executable target set $\mathcal{T}_{\mathrm{feas}}(t_k)$; occupancy indicators $\{q_{i,k}\}_{i=1}^{K_{\max}}$; completion indicators $\{c_{i,k}\}_{i=1}^{K_{\max}}$; communication availability $I_{\mathrm{comm}}(t_k)$; onboard storage $S(t_k)$; resource-safe action set $\mathcal{A}_{\mathrm{safe}}(t_k)$
\State \textbf{Output.} Mask vector $\bm{m}_k\in\{0,1\}^{K_{\max}+2}$; admissible action set $\mathcal{A}_{\mathrm{adm}}(s_k)$

\State Initialize the mask vector $\bm{m}_k\leftarrow\mathbf{0}_{K_{\max}+2}$
\For{$i=1,2,\dots,K_{\max}$}
    \If{$q_{i,k}=1$}
        \If{$\tau_i\in\mathcal{T}_{\mathrm{feas}}(t_k)$}
            \If{$c_{i,k}=0$}
                \If{$a_i^{\mathrm{obs}}\in\mathcal{A}_{\mathrm{safe}}(t_k)$}
                    \State Set $m_{i,k}^{\mathrm{obs}}\leftarrow 1$
                \EndIf
            \EndIf
        \EndIf
    \EndIf
\EndFor

\If{$I_{\mathrm{comm}}(t_k)=1$}
    \If{$S(t_k)>0$}
        \If{$a^{\mathrm{dl}}\in\mathcal{A}_{\mathrm{safe}}(t_k)$}
            \State Set $m_{K_{\max}+1,k}^{\mathrm{dl}}\leftarrow 1$
        \EndIf
    \EndIf
\EndIf

\If{$a^{\mathrm{idle}}\in\mathcal{A}_{\mathrm{safe}}(t_k)$}
    \State Set $m_{K_{\max}+2,k}^{\mathrm{idle}}\leftarrow 1$
\EndIf

\If{$\sum_{j=1}^{K_{\max}+2}m_{j,k}=0$}
    \State Set $m_{K_{\max}+2,k}^{\mathrm{idle}}\leftarrow 1$ as an idle fallback to guarantee a non-empty admissible action set
\EndIf

\State Construct the admissible action set
\[
\mathcal{A}_{\mathrm{adm}}(s_k)
=
\left\{
a_j\in\mathcal{A}
\mid
m_{j,k}=1
\right\}
\]

\State \Return $\bm{m}_k$, $\mathcal{A}_{\mathrm{adm}}(s_k)$
\end{algorithmic}
\end{algorithm}

Based on the physics-constrained maskable policy, the PPO probability ratio is written as
\begin{equation}
\rho_k(\theta)
=
\frac{\tilde{\pi}_{\theta}(a_k\mid s_k,\bm{m}_k)}
{\tilde{\pi}_{\theta_{\mathrm{old}}}(a_k\mid s_k,\bm{m}_k)}.
\end{equation}
The clipped surrogate objective then becomes
\begin{equation}
L^{\mathrm{clip}}(\theta)
=
\mathbb{E}_k
\left[
\min
\left(
\rho_k(\theta)\hat{A}_k,\;
\mathrm{clip}\!\left(\rho_k(\theta),1-\epsilon,1+\epsilon\right)\hat{A}_k
\right)
\right],
\end{equation}
where $\epsilon$ is the clipping threshold. The value-function loss and the entropy regularization term are defined as
\begin{equation}
\begin{aligned}
L^{\mathrm{VF}}(\phi)
&=
\mathbb{E}_k
\left[
\big(
V_{\phi}(s_k)-\hat{R}_k
\big)^2
\right], \\
L^{\mathrm{ent}}(\theta)
&=
\mathbb{E}_k
\left[
\mathcal{H}\big(
\tilde{\pi}_{\theta}(\cdot\mid s_k,\bm{m}_k)
\big)
\right].
\end{aligned}
\end{equation}
where $\hat{R}_k=\hat{A}_k+V_{\phi}(s_k) $ denotes the return target. The overall optimization objective is therefore given by
\begin{equation}
L(\theta,\phi)
=
L^{\mathrm{clip}}(\theta)
-
c_{\mathrm{v}} L^{\mathrm{VF}}(\phi)
+
c_{\mathrm{e}} L^{\mathrm{ent}}(\theta),
\end{equation}
where $c_{\mathrm{v}}$ and $c_{\mathrm{e}}$ are the weighting coefficients for the value loss and entropy regularization, respectively. In implementation, the negative of $L(\theta,\phi)$ is minimized using Adam.
By embedding the physics-constrained mask into the policy distribution, feasible action selection is enforced at the policy level rather than corrected only after action execution. Consequently, the PCM-PPO update is restricted to the admissible action set determined by the current mission state, so that the agent allocates probability only to executable actions and updates its policy only over the feasible decision domain.

\subsection{Network structure and training process}\label{3.3}

\begin{figure*}[!t]
  \centering
  \includegraphics[width=16cm]{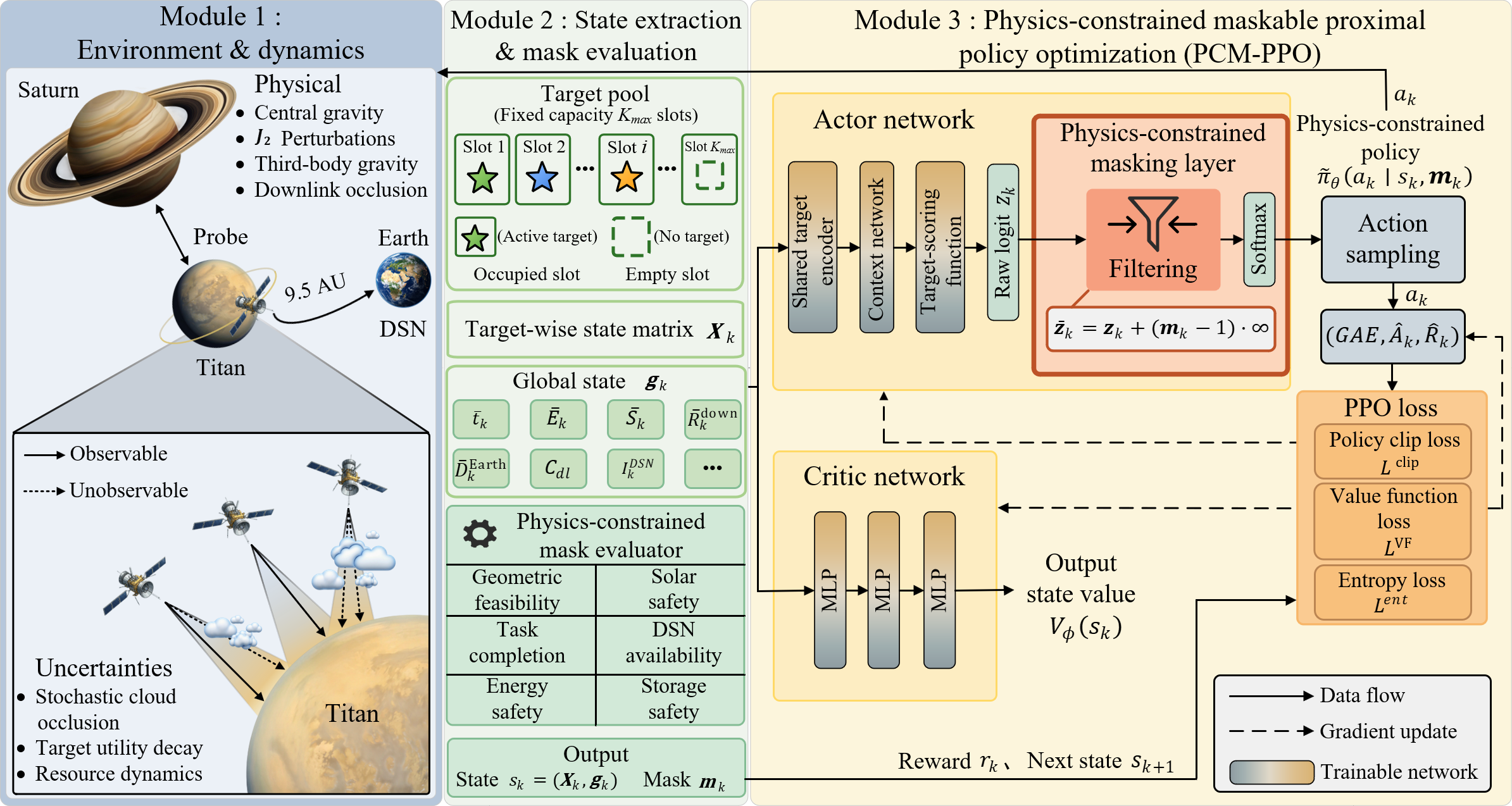}
  \caption{Overall framework of the PCM-PPO method.}
  \label{fig_3.2}
\end{figure*}

The overall framework of the proposed PCM-PPO method is illustrated in \figref{fig_3.2}. The framework consists of three coupled modules, namely the Titan mission environment and dynamics module, the state extraction and mask evaluation module, and the physics-constrained maskable proximal policy optimization module. In Module 1, the Titan mission environment is modeled by considering multi-body orbital dynamics, geometric observability, downlink occultation, and stochastic mission uncertainties, including cloud occlusion, target utility decay, resource dynamics, and dynamic target arrival. These factors jointly determine the time-varying mission state and the feasibility of observation and downlink operations.

Module 2 converts the current mission condition into a structured state representation and a physics-constrained action mask. Different from a conventional fixed-target state vector, the target set is represented by a fixed-capacity target pool with $K_{\max}$ slots. Each slot can either be occupied by an active target or remain empty. Newly arrived opportunity targets are inserted into empty slots after their arrival, while empty slots are excluded from action selection through the occupancy indicator. Based on the target pool, a target-wise state matrix $\bm{X}_k$ is constructed to describe the slot-level target attributes, including target occupancy, utility, cloud probability, geometric feasibility, solar-safety status, visibility realization, completion flag, and maneuver demand. Meanwhile, the global mission-state vector $\bm{g}_k$ describes the mission progress, onboard energy, storage state, downlink condition, Earth-probe distance, and DSN availability. The resulting state is represented as $s_k=(\bm{X}_k,\bm{g}_k)$. In parallel, the physics-constrained mask evaluator constructs the mask vector $\bm{m}_k$ according to slot occupancy, geometric feasibility, solar safety, cloud visibility, task completion, communication availability, and resource safety. Thus, the mask is derived from the Titan mission state rather than being treated as an externally specified validity signal.

Module 3 performs policy optimization based on the structured state and the physics-constrained mask. The actor network adopts a target-conditioned architecture. Specifically, the shared target encoder first maps each target-slot feature into a latent representation. The context network then aggregates the active target representations and combines them with the global mission state to obtain a mission context. Based on this context, the target-scoring function generates raw logits for target-observation actions, while the downlink and idle decisions are represented in the same action space. Before the softmax operation, the physics-constrained masking layer filters the raw logits by assigning infeasible actions and empty target slots zero probability in the resulting policy distribution. Therefore, action sampling and subsequent PPO updates are performed over the physically admissible decision domain. The critic network estimates the state value for policy evaluation, and the PPO objective jointly considers the clipped policy loss, value-function loss, and entropy regularization term.

\begin{table*}[t]
\centering
\caption{Functional comparison of representative planning and reinforcement-learning formulations for Titan autonomous mission planning.}
\label{tab:functional_comparison}
\small 
\setlength{\tabcolsep}{4pt} 
\renewcommand{\arraystretch}{1.12}
\begin{tabular}{lcccccc}
\hline
Formulation
& \makecell[c]{Dynamic targets}
& \makecell[c]{Fixed action size}
& \makecell[c]{Feasible\\action constraint}& \makecell[c]{Coupled\\constraints}
& \makecell[c]{Uncertainty response}
& \makecell[c]{Low online cost} \\
\hline
Standard PPO
& $\triangle$ & $\checkmark$ & $\times$ & $\triangle$ & $\checkmark$ & $\checkmark$ \\
Safe / Constrained RL
& $\triangle$ & $\checkmark$ & $\checkmark$ & $\triangle$ & $\checkmark$ & $\checkmark$ \\
Receding-horizon optimization
& $\checkmark$ & $\triangle$& $\checkmark$ & $\checkmark$ & $\checkmark$ & $\triangle$ \\
Metaheuristic scheduling
& $\triangle$ & $\triangle$ & $\times$ & $\triangle$ & $\triangle$ & $\times$ \\
PCM-PPO
& $\checkmark$ & $\checkmark$ & $\checkmark$ & $\checkmark$ & $\checkmark$ & $\checkmark$ \\
\hline
\multicolumn{7}{p{0.98\textwidth}}{\footnotesize
$\checkmark$: supported;
$\triangle$: supported only with additional problem-specific design or repeated re-planning;
$\times$: not naturally supported.}
\end{tabular}
\end{table*}

In terms of network implementation, the actor adopts a target-conditioned architecture rather than a conventional fixed-index multilayer perceptron. The target-feature matrix $\bm{X}_k$ is first processed by the shared target encoder $g_{\psi}(\cdot)$, which maps each target slot into a latent target embedding $\bm{h}_{i,k}$. Since the same encoder is applied to all target slots, the policy evaluates targets according to their current physical and mission attributes rather than relying on fixed target indices. This design is particularly suitable for the bounded dynamic target pool, because newly arrived opportunity targets can be inserted into empty slots and evaluated by the same representation and scoring mechanism as predefined targets.

The active target embeddings are then aggregated and combined with the global mission-state vector $\bm{g}_k$ through the context network $\rho(\cdot)$ to obtain the set-level mission context $\bm{c}_k$. Based on this context, the shared target-scoring function $f_{\theta}(\cdot)$ generates logits for all target-observation actions, while two additional action heads generate the logits for downlink and idle actions. The physics-constrained masking layer is subsequently applied to the resulting logits so that inactive target slots and physically infeasible actions receive zero probability after the softmax operation.

The critic network estimates the scalar state value from the structured mission state for policy evaluation. Therefore, the target-conditioned actor is used to generate the masked policy distribution, while the critic provides value estimates for advantage estimation and PPO updating. Compared with a standard maskable PPO implementation that directly maps a flattened state vector to action logits, the proposed architecture separates target representation, mission-context extraction, target-action scoring, and physics-constrained feasibility filtering, thereby providing a more structured policy model for Titan mission planning.
To further clarify the methodological distinction, \tabref{tab:functional_comparison} compares PCM-PPO with representative planning and reinforcement-learning formulations in terms of dynamic target handling, action dimensionality, feasibility constraints, coupled constraints, uncertainty response, and online decision cost.

To improve training stability and convergence under high dimensional state representations and strongly coupled physical and operational constraints, an integrated training strategy is introduced in the PCM-PPO method. First, advantage normalization is applied to the generalized advantage estimation (GAE) output. Specifically, the estimated advantage is normalized as
\begin{equation}
\tilde{A}_k=\frac{\hat{A}_k-\mu_A}{\sigma_A+\varepsilon_A},
\end{equation}
where $\hat{A}_k$ denotes the GAE-based advantage estimate at decision step $k$, $\mu_A$ and $\sigma_A$ denote the mean and standard deviation of the advantage values within the current training batch, respectively, and $\varepsilon_A$ is a small positive constant introduced for numerical stability. Advantage normalization reduces the influence of inter-batch fluctuations in the advantage scale on policy-gradient updates, thereby improving the stability of the training process.
We also employ orthogonal initialization for both the policy network and the value network to alleviate scale imbalance in the early stage of forward propagation. Constraining the initial parameter scale to a more balanced range, orthogonal initialization facilitates maintaining stable signal propagation across layers and mitigates excessively aggressive parameter updates at the beginning of training. The network parameters are optimized using the Adam optimizer, which provides robust adaptive updates under noisy gradients and long-horizon return estimation. Gradient clipping is further applied during parameter optimization to suppress update oscillations and improve training stability by preventing excessively large gradient steps.
Simultaneously, to balance global exploration in the early stage of training with refined convergence in the later stage, a learning-rate decay is introduced and written as
\begin{equation}
\eta(p)=\eta_{\min}+(\eta_0-\eta_{\min})p,
\end{equation}
where $p$ denotes the remaining training progress, $\eta_0$ is the initial learning rate, and $\eta_{\min}$ is the lower-bound learning rate used near the end of training. Consequently, our model maintains a relatively larger learning rate during the early stage to enhance exploration efficiency, while a gradually reduced learning rate is adopted in the later stage to suppress parameter oscillation and improve convergence precision. 

\begin{algorithm}[t]
\caption{PCM-PPO training procedure}
\label{alg:pcmppo_full}
\begin{algorithmic}[1]
\State \textbf{Input.} Mission environment $\mathcal{E}$; policy parameters $\theta$; value parameters $\phi$; discount factor $\gamma$; GAE parameter $\lambda$; clipping threshold $\epsilon$; coefficients $c_{\mathrm{v}}$ and $c_{\mathrm{e}}$; learning rates $\eta_0$ and $\eta_{\min}$; rollout length $T_{\mathrm{roll}}$; training episodes $N_{\mathrm{ep}}$; update epochs $K$; mini-batch size $B$
\State \textbf{Output.} Optimized policy $\theta^{\ast}$ and value parameters $\phi^{\ast}$
\State Initialize the target encoder $g_{\psi}$, context network $\rho$, action heads, and critic $V_{\phi}$
\For{$e=1,2,\ldots,N_{\mathrm{ep}}$}
    \State Reset $\mathcal{E}$, initialize the target pool, and obtain $s_0=(\bm{X}_0,\bm{g}_0)$
    \State Initialize rollout buffer $\mathcal{D}\leftarrow\emptyset$
    \For{$k=0,1,\ldots,T_{\mathrm{roll}}-1$}
        \State Update the target pool with newly arrived opportunity targets
        \State Construct $\bm{m}_k$ and $\mathcal{A}_{\mathrm{adm}}(s_k)$ using Algorithm~\ref{alg:mask_full}
        \State Generate target-conditioned logits $\bm{z}_k$ using $g_{\psi}$, $\rho$, and action heads
        \State Apply $\bm{m}_k$ to obtain $\bar{\bm{z}}_k$ and masked policy $\tilde{\pi}_{\theta}(\cdot\mid s_k,\bm{m}_k)$
        \State Sample $a_k\sim\tilde{\pi}_{\theta}(\cdot\mid s_k,\bm{m}_k)$
        \State Execute $a_k$, observe $r_k$, $s_{k+1}$, and terminal flag $d_k$
        \State Store $(s_k,\bm{m}_k,a_k,r_k,s_{k+1},d_k)$ in $\mathcal{D}$
        \If{$d_k=1$}
            \State \textbf{break}
        \EndIf
    \EndFor

    \State Compute TD residuals $\delta_k$, GAE advantages $\hat{A}_k$, and returns $\hat{R}_k$ for all transitions in $\mathcal{D}$
    \State Normalize advantages to obtain $\tilde{A}_k$
    \State Update learning rate by $\eta(p)=\eta_{\min}+(\eta_0-\eta_{\min})p$, where $p=1-e/N_{\mathrm{ep}}$
    \State Set $\theta_{\mathrm{old}}\leftarrow\theta$
    \For{$u=1,2,\ldots,K$}
        \State Sample a mini-batch $\mathcal{B}\subset\mathcal{D}$ of size $B$
        \State Compute the probability ratio $\rho_k(\theta)$ for each sample in $\mathcal{B}$
        \State Compute $L^{\mathrm{clip}}(\theta)$, $L^{\mathrm{VF}}(\phi)$, $L^{\mathrm{ent}}(\theta)$, and $L(\theta,\phi)$
        \State Update $\theta$ and $\phi$ using Adam with gradient clipping
    \EndFor
\EndFor
\State \Return $\theta^{\ast}\leftarrow\theta$, $\phi^{\ast}\leftarrow\phi$
\end{algorithmic}
\end{algorithm}

The complete algorithmic implementation and training procedure of the PCM-PPO method are summarized in Algorithm~\ref{alg:pcmppo_full}.

\begin{figure*}[!t]
  \centering
  \includegraphics[width=16cm]{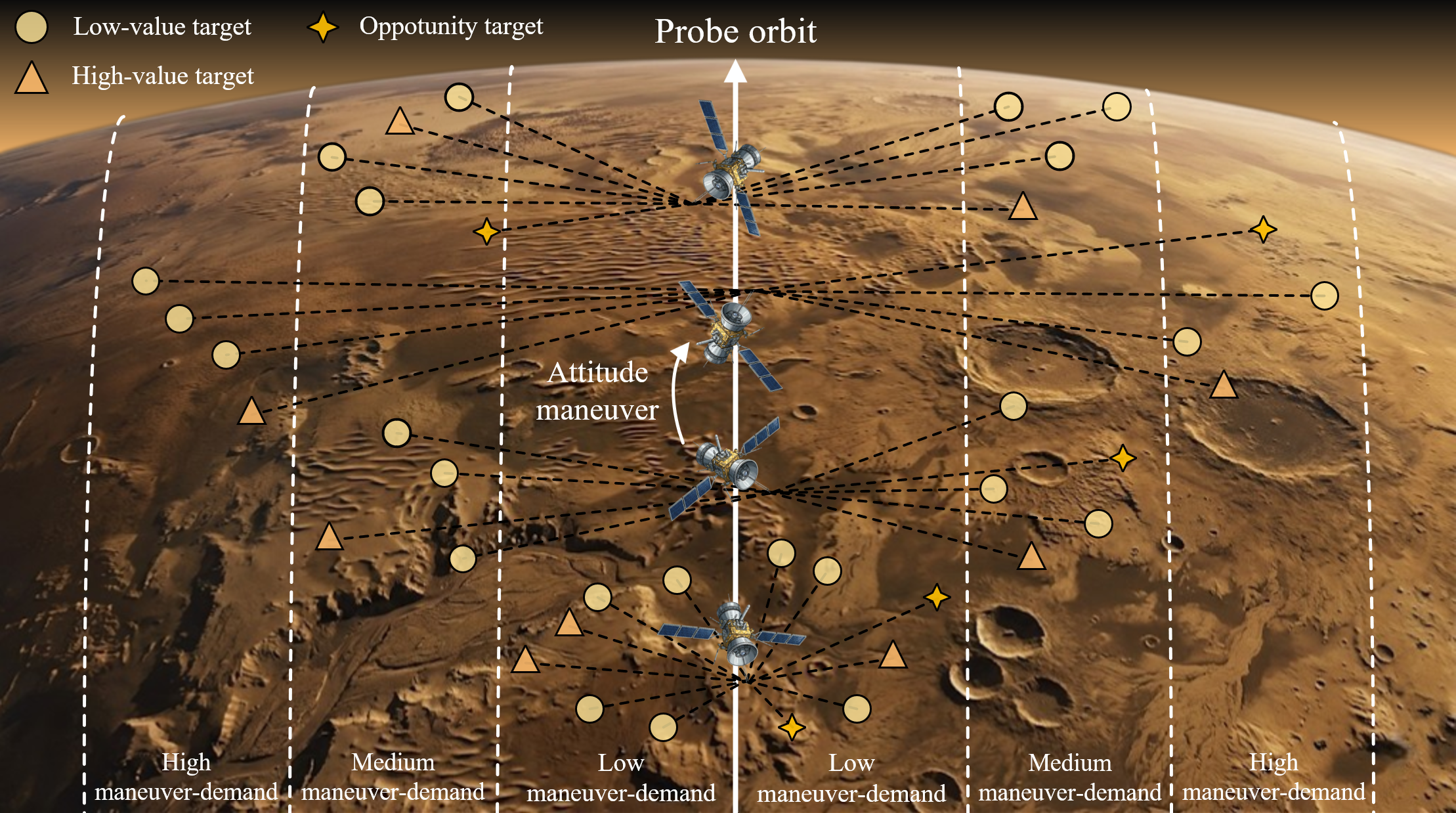}
  \caption{Schematic illustration of probe attitude maneuvers for targets in different maneuver-demand bands}
  \label{fig_3.3}
\end{figure*}

In addition, to improve the physical interpretability of the mission scenarios under perturbed Titan-centered orbits, the observation geometry of candidate targets is characterized according to the required attitude maneuver demand rather than absolute latitude \cite{elachi2005cassini,lorenz2021selection}. 
Under $J_2$ perturbation and Saturn's third-body attraction, the probe-target relative observation geometry varies over time, and thus the maneuver effort required for target observation cannot be directly represented by the target's absolute latitude and surface distance. Instead, a maneuver-demand-based characterization mechanism is introduced, in which a normalized maneuver-demand index is constructed for each candidate target $\tau_i$ at time $t$ as
\begin{equation}
\eta_i(t)=
\max\left(
\frac{|\theta_{\mathrm{lr},i}(t)|}{\alpha_{\mathrm{lr}}},
\frac{|\theta_{\mathrm{fb},i}(t)|}{\alpha_{\mathrm{fb}}}
\right),
\end{equation}
where $\theta_{\mathrm{lr},i}(t)$ and $\theta_{\mathrm{fb},i}(t)$ denote the cross-track and along-track viewing angles of target $\tau_i$, respectively, and $\alpha_{\mathrm{lr}}$ and $\alpha_{\mathrm{fb}}$ are the corresponding allowable field-of-view limits. A larger $\eta_i(t)$ indicates that the target lies farther away from the nominal viewing direction and requires a larger off-nadir attitude maneuver, as illustrated in \figref{fig_3.3}.

Based on this index, candidate targets are further divided into different maneuver-demand bands to represent different levels of maneuver-related resource consumption.
\begin{enumerate}
    \item Low-demand region: $0 \leq \eta_i < 1/3$, representing targets that remain close to the nominal viewing direction and can be observed with little or no additional attitude maneuver.
    \item Medium-demand region: $1/3 \leq \eta_i < 2/3$, indicating targets that require a moderate off-nadir attitude maneuver and therefore incur additional maneuver-related resource consumption.
    \item High-demand region: $2/3 \leq \eta_i \leq 1$, corresponding to targets located close to the edge of the feasible viewing envelope and requiring relatively large off-nadir attitude maneuvers, thus leading to the highest maneuver-related resource consumption, which is reflected in the action energy term $e(a_t)$.
\end{enumerate}

Within this maneuver-demand-based grouping, candidate targets are further categorized as low-value targets, high-value targets, and opportunity targets according to their scientific utility. Low-value and high-value targets are used to represent routine observations with different reward levels, whereas opportunity targets describe transient events with stronger temporal urgency and are assigned higher initial utility together with temporal decay. Consequently, the target distribution is coupled not with heuristic latitude bands, but with the physically meaningful observation geometry and utility-driven mission priorities. This mechanism improves the interpretability of training scenarios and enables a better balance between scientific return and maneuver-related resource consumption in multi-target autonomous mission planning.

\section{Experiments}\label{4}

\subsection{Mission scenario and parameter setting}\label{4.1}

To evaluate the proposed PCM-PPO method, experiments were conducted in the Titan mission-planning environment established in \secref{2} and \secref{3}. The simulation scenario was constructed in a Titan-centered, parameterized, and seed-controlled manner so that all compared methods were evaluated under the same physical, geometric, resource, and uncertainty conditions. The main physical and simulation parameters are summarized in \tabref{tab:physical_params}. In the experiments, the probe orbital altitude was set to 1500 km, the decision step size was $\Delta t=60$ s, and each episode contained 1440 decision steps, corresponding to a 24 h planning horizon. The field-of-view limits, solar-elevation threshold, sun-avoidance angle, minimum DSN elevation angle, cloud-model parameters, utility-decay parameters, energy-storage parameters, downlink parameters, and safety margins followed the settings listed in \tabref{tab:physical_params}.

\begin{table}[!t]
\centering
\caption{Physical and simulation parameters of the Titan mission environment.}
\label{tab:physical_params}
\footnotesize
\setlength{\tabcolsep}{3pt}
\renewcommand{\arraystretch}{1.08}
\begin{tabularx}{\columnwidth}{@{}>{\raggedright\arraybackslash}X c >{\raggedright\arraybackslash}p{0.30\columnwidth}@{}}
\hline
Parameter & Symbol & Value \\
\hline
Titan gravitational parameter & $\mu_{\mathrm{T}}$ & $8978.138\ \mathrm{km^3/s^2}$ \\
Titan mean radius & $R_{\mathrm{T}}$ & $2575.0\ \mathrm{km}$ \\
Saturn mean radius & $R_{\mathrm{S}}$ & $60268\ \mathrm{km}$ \\
Titan zonal harmonic coefficient & $J_{2,\mathrm{T}}$ & $31.81\times10^{-6}$ \\
Titan rotation rate & $\omega_{\mathrm{T}}$ & $4.560\times10^{-6}\ \mathrm{rad/s}$ \\
Saturn gravitational parameter & $\mu_{\mathrm{S}}$ & $3.79312\times10^{7}\ \mathrm{km^3/s^2}$ \\
Titan--Saturn distance & $D_{\mathrm{TS}}$ & $1\,221\,870\ \mathrm{km}$ \\
Reference Earth--probe distance & $D_{\oplus,\mathrm{ref}}$ & $9.5\ \mathrm{AU}$ \\
Nominal one-way communication delay & $t_{\mathrm{comm}}$ & $\approx 79\ \mathrm{min}$ \\
Probe orbital altitude & $h$ & $1500\ \mathrm{km}$ \\
Decision step size & $\Delta t$ & $60\ \mathrm{s}$ \\
Planning horizon per episode & $M$ & $1440$ steps ($24$ h) \\
Cross-track FOV limit & $\alpha_{\mathrm{lr}}$ & $28.1^\circ$ \\
Along-track FOV limit & $\alpha_{\mathrm{fb}}$ & $25.0^\circ$ \\
Minimum solar elevation angle & $\theta_{\odot,\min}$ & $5.0^\circ$ \\
Sun-avoidance angle & $\phi_{\mathrm{safe}}$ & $25.0^\circ$ \\
Minimum DSN elevation angle & $\epsilon_{\min}$ & $10.0^\circ$ \\
DSN contact window length & $|W_{\mathrm{DSN}}|$ & $8$ h \\
\hline
Target-slot cloud-free prior & $\bar{\varpi}_i$ & $[0.2,0.5]$ \\
Cloud persistence coefficient & $\mu_{\mathrm{c}}$ & $0.85$ \\
Cloud update noise standard deviation & $\sigma_{\mathrm{c}}$ & $0.1$ \\
Cloud update reference interval & $\Delta t_{\mathrm{c}}$ & $60\ \mathrm{s}$ \\
Minimum executable target utility & $U_{\min}$ & $10$ \\
Initial utility levels & $U^{\mathrm{init}}_{\mathrm{low/high/opp}}$ & $30/60/90$ \\
Utility decay rate of predefined targets & $\lambda_{\mathrm{base}}$ & $0$ \\
Utility decay rate of opportunity targets & $\lambda_{\mathrm{opp}}$ & $2$--$5\ \mathrm{day^{-1}}$ \\
RTG power generation & $P_{\mathrm{RTG}}$ & $500\ \mathrm{W}$ \\
Platform baseline power & $P_{\mathrm{base}}$ & $450\ \mathrm{W}$ \\
Observation payload power & $P_{\mathrm{obs}}$ & $500\ \mathrm{W}$ \\
Communication power & $P_{\mathrm{comm}}$ & $500\ \mathrm{W}$ \\
Onboard storage capacity & $S_{\max}$ & $20\ \mathrm{Gbit}$ \\
Onboard energy capacity & $E_{\max}$ & $2000\ \mathrm{Wh}$ \\
Baseline data volume per observation & $d_{\mathrm{obs}}$ & $1000\ \mathrm{Mbit}$ \\
Reference downlink rate & $R_{\mathrm{ref}}$ & $1200\ \mathrm{kbps}$ \\
Energy safety margin & $\Delta E_{\mathrm{safe}}$ & $200\ \mathrm{Wh}$ \\
Training seeds & $s_{\mathrm{rand}}$ & $1/2/3/4/5$ \\
Independent Monte Carlo trials & $N_{\mathrm{MC}}$ & $20$ \\
\hline
\end{tabularx}
\end{table}

\begin{table}[htbp]
\centering
\caption{Training hyperparameters, network settings, and reward coefficients of the PCM-PPO model.}
\label{tab:training_params}
\renewcommand{\arraystretch}{1.12}
\setlength{\tabcolsep}{3pt}
\small
\begin{tabular}{p{0.42\columnwidth}p{0.28\columnwidth}p{0.22\columnwidth}}
\hline
Hyperparameter & Symbol / Setting & Value \\
\hline
Target feature dimension & $d_x$ & $8$ \\
Global state dimension & $d_g$ & $7$ \\
Target encoder hidden layers & $g_{\psi}$ & $[256,256]$ \\
Target embedding dimension & $\bm{h}_{i,k}$& $256$ \\
Context network hidden layers & $\rho$ & $[256,256]$ \\
Mission context dimension & $\bm{c}_k$& $256$ \\
Target-scoring head & $f_{\theta}$ & $[256,1]$ \\
Downlink/idle action heads & $f_{\theta}^{\mathrm{dl}}$, $f_{\theta}^{\mathrm{idle}}$ & $[256,1]$ \\
Critic hidden layers & $V_{\phi}$ & {\fontsize{8}{9.6}\selectfont$[256,256,256]$}\\
Activation function & --- & Tanh \\
Optimizer & --- & Adam \\
Total training timesteps & --- & $1.0\times10^6$ \\
Rollout/episode length & $n_{\mathrm{steps}}$ & $1440$ \\
Planning step size & $\Delta t$ & $60\ \mathrm{s}$ \\
Batch size & $B$ & $144$ \\
Training epochs per update & $K$ & $10$ \\
Discount factor & $\gamma$ & $0.99$ \\
Entropy coefficient & $c_{\mathrm{e}}$ & $0.01$ \\
Learning-rate schedule & $\eta(p)$ & $3\times10^{-4}$ to $1\times10^{-4}$ \\
\hline
Utility normalization constant & $U_{\max}$ & $100$ \\
Observation reward coefficient & $\lambda_{\mathrm{obs}}$ & $100$ \\
Downlink reward coefficient & $\lambda_{\mathrm{dl}}$ & $500$ \\
Energy-consumption penalty coefficient & $\lambda_E$ & $100$ \\
Storage-growth penalty coefficient & $\lambda_S$ & $25$ \\
Idle penalty coefficient & $\lambda_{\mathrm{idle}}$ & $0.05$ \\
\hline
\end{tabular}
\end{table}

The Titan-centered geometric scenario was generated as follows. The probe orbit was propagated around Titan using the dynamics model in \secref{2.1}. Surface targets were sampled in Titan-fixed latitude--longitude coordinates and then transformed into the Titan-centered inertial frame using Titan's rotation model. The Titan-Saturn distance and Saturn radius in \tabref{tab:physical_params} were used to evaluate Saturn-related geometric occultation and safety constraints, while the Earth and Sun vector directions were used for communication-line and illumination checks. Therefore, the generated scenario preserves the main planning-level geometric couplings among probe motion, target observability, solar illumination, and Earth downlink feasibility.

For DSN availability, each 24 h episode contained one scheduled contact window with duration $|W_{\rm DSN}|=8$ h. At the beginning of each daily scenario, the contact-window start phase $t_{\rm DSN}^{0}$ was sampled from $[0,T_{\rm day}-|W_{\rm DSN}|]$ using the scenario seed, and the corresponding contact window was defined as $W_{\rm DSN}=[t_{\rm DSN}^{0},t_{\rm DSN}^{0}+|W_{\rm DSN}|)$. The DSN-window indicator was then computed as $I_k^{\rm DSN}=\mathbf{I}(t_k\in W_{\rm DSN})$. This scheduled window represents ground-network availability only. A downlink action was executable only when the current decision time was inside $W_{\rm DSN}$ and the physical communication feasibility $C_{\rm dl}(t_k)$, onboard-storage condition, and resource-safety condition were simultaneously satisfied.

For the mission target design, each Monte Carlo scenario contained predefined mapping targets and dynamically arriving opportunity targets. The target categories followed a fixed proportion of 50\%, 30\%, and 20\% for low-value targets, high-value targets, and opportunity targets, respectively. Target locations were generated by Monte Carlo sampling in Titan-fixed latitude-longitude coordinates, and the resulting targets were organized according to the maneuver-demand-based characterization mechanism described in \secref{3.3}. In the present setting, opportunity-target arrival times and associated attributes are generated by the Monte Carlo scenario environment rather than by an onboard target-discovery, detection, or identification module. Specifically, the Titan-fixed location is sampled as described above, whereas the initial utility, cloud-free probability, and utility-decay rate are assigned according to the target category and the parameter settings in \tabref{tab:physical_params}; the completion flag is initialized to zero. Predefined mapping targets were initialized at the beginning of each episode and inserted into the fixed-capacity target pool described in \secref{2.2}. In contrast, opportunity targets were assigned arrival times within the 24 h scenario and remained inactive before arrival. Before an opportunity target arrived, its corresponding target slot had $q_{i,k}=0$, its target-level features were set to zero, and its associated observation action was excluded by the physics-constrained mask. Once the opportunity target arrived, its Titan-fixed location, initial utility, cloud-free probability, utility-decay rate, and completion flag were instantiated and inserted into an available opportunity-target slot. Therefore, opportunity targets were not available to the policy before arrival, and the agent could not score or select the corresponding observation actions until they entered the fixed-capacity target pool.
The utility-decay rates in \tabref{tab:physical_params} were expressed in day$^{-1}$. For opportunity targets, $\lambda_{\rm opp}$ was sampled from $[2,5]$ day$^{-1}$, corresponding to an approximate utility half-life of 3.3-8.3 h. This setting makes opportunity targets time-sensitive within the daily planning horizon while avoiding an unrealistically instantaneous disappearance at the 60 s decision-step scale. In the utility update, the elapsed time in the exponential decay term was converted to days to maintain consistency with the unit of $\lambda_{\rm opp}$.

The training hyperparameters, network settings, and reward coefficients of PCM-PPO are summarized in \tabref{tab:training_params}. The actor network was implemented using the target-conditioned structure described in \secref{3.3}. The shared target encoder and context network both used two hidden layers with 256 neurons, and the target-scoring, downlink, and idle action heads generated the corresponding action logits. The critic network used three hidden layers with 256 neurons to estimate the scalar state value. 
Tanh was used as the activation function, and Adam was adopted as the optimizer. The total training timesteps were set to $1.0\times10^6$. The rollout length was 1440, the batch size was 144, and the number of training epochs per update was 10. The discount factor was set to 0.99, the entropy coefficient was 0.01, and the learning rate was initialized at $3\times10^{-4}$ and decayed to $1\times10^{-4}$ during training. The reward coefficients are used to define a normalized scalar planning objective rather than calibrated physical science values. The observation term is kept dominant to prioritize scientific return, the downlink term encourages data return and storage relief, and the resource terms act as shaping penalties under the hard resource-safety mask.

All experiments were implemented in Python 3.10 with PyTorch and conducted on a Windows 10 workstation equipped with an NVIDIA GeForce RTX 3060 GPU, a 12th Gen Intel(R) Core(TM) i7-12700 CPU (2.10 GHz), and 64.0 GB RAM. For subsequent experiments, all methods were evaluated under the same simulation environment, planning-horizon setting, target-generation seeds, and Monte Carlo scenario seeds.

In addition to the training and evaluation settings, \tabref{tab:deployment} reports the deployment-related computational characteristics of PCM-PPO measured on the experimental CPU workstation. The online decision process consists of neural-network inference and physics-constrained mask evaluation. If the neural-network output is abnormal, the mask evaluation fails, the admissible action set becomes empty, or the selected action is found infeasible, the planner switches to an idle or standby fallback to maintain a conservative executable action.

\begin{table}[!t]
\centering
\caption{Deployment-related computational characteristics of PCM-PPO.}
\label{tab:deployment}
\renewcommand{\arraystretch}{1.12}
\setlength{\tabcolsep}{4pt}
\footnotesize
\begin{tabular}{p{0.43\linewidth} p{0.49\linewidth}}
\hline
Item & Value \\
\hline
Trainable parameters 
& Full actor--critic: 0.335--0.434 M; actor-only deployment path: 0.170--0.224 M \\
Model size, FP32 / FP16
& Full model: 1.28--1.66 MB / 0.64--0.83 MB; actor-only path: 0.65--0.85 MB / 0.32--0.43 MB\\
Mask construction time
& 0.92--2.01 ms mean; $\leq$2.52 ms at $p_{95}$ \\
Actor inference time
& 0.39--0.42 ms mean; $\leq$0.55 ms at $p_{95}$ \\
Total online decision time
& 1.32--2.43 ms mean; $\leq$2.97 ms at $p_{95}$ \\
Online computation type
& Neural network inference and mask evaluation\\
Safety fallback mechanism
& Idle / standby fallback \\
\hline
\end{tabular}
\end{table}

\subsection{Training performance and convergence }\label{4.2}
\begin{figure}[!t]
  \centering
  \includegraphics[width=7.5cm]{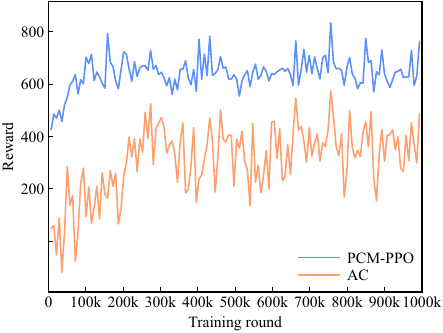}
  \caption{Comparison of training reward curves between PCM-PPO and AC.}
  \label{fig_4_1}
\end{figure}

\begin{figure}[!t]
  \centering
  \includegraphics[width=7.5cm]{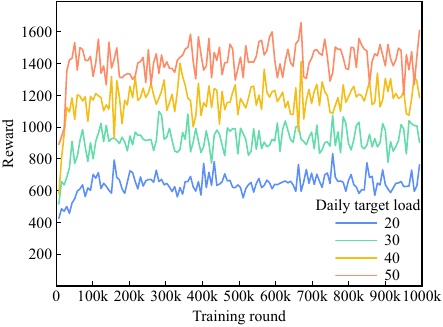}
  \caption{Training reward curves of PCM-PPO under varying daily target loads.}
  \label{fig_4_2}
\end{figure}

To assess the training performance of PCM-PPO, the reward evolution during the training process is presented in \figref{fig_4_1} and \figref{fig_4_2}. The reward value is recorded at a fixed evaluation interval during training and represents the cumulative daily reward over one $24$h planning horizon under a given daily target load.

\figref{fig_4_1} compares the training reward curves of PCM-PPO and the AC baseline over the full training process. As depicted in the figure, PCM-PPO achieves a markedly faster reward increase in the early stage and maintains a consistently higher reward level throughout training. After the initial rapid growth phase, the PCM-PPO curve enters a relatively stable regime and fluctuates within a bounded interval, indicating that the learned policy gradually reaches stable planning performance under the coupled mission constraints. By contrast, although the AC baseline also improves during training, its reward remains substantially lower and exhibits stronger oscillations, suggesting PCM-PPO consistently outperforms AC.

The convergence performance of PCM-PPO under varying daily target loads is further illustrated in \figref{fig_4_2}. For all target loads, i.e., daily target loads of 20, 30, 40, and 50, the reward increases rapidly at the beginning of training and then gradually approaches a stable range, while the converged reward level also increases with the daily target load. Moreover, the late-stage reward curves remain stable under all target load settings, indicating that PCM-PPO preserves favorable convergence performance as the planning target load increases.

\subsection{Comparative performance under varying target loads}\label{4.3}

\begin{table*}[t]
\centering
\caption{Baseline configurations used in the comparative experiments.}
\label{tab:baseline_settings}
\scriptsize
\setlength{\tabcolsep}{3.2pt}
\renewcommand{\arraystretch}{1.06}
\begin{tabular}{@{}p{1.2cm}p{2.05cm}p{1.9cm}p{3.8cm}p{5.5cm}p{1cm}@{}}
\hline
Method & Type & Model / encoding & Budget & Key parameters & Constraint \\
\hline
PB-PPO
& Policy-based RL
& Actor--critic MLP, $4\times256$, Tanh
& $1.0\times10^{6}$ steps
& $\gamma=0.99$; clip $=0.2$; GAE $\lambda=0.95$; $c_{\mathrm{e}}=0.01$; actor lr $=2\times10^{-4}$; critic lr $=7\times10^{-4}$; epochs $=4$
& Penalty \\
PPO
& Policy-based RL
& Actor--critic MLP, $3\times256$, ReLU
& $1.0\times10^{6}$ steps
& $\gamma=0.99$; clip $=0.2$; GAE $\lambda=0.95$; $c_{\mathrm{e}}=0.01$; batch size $=144$
& Penalty \\
AC
& Actor--critic RL
& Actor--critic MLP, $3\times256$, ReLU
& $1.0\times10^{6}$ steps
& $\gamma=0.99$; $c_{\mathrm{e}}=0.01$; value-loss coefficient $=0.5$; learning rate $3\times10^{-4}$ to $1\times10^{-4}$
& Penalty \\
DQN
& Value-based RL
& Q-network MLP, $3\times256$, ReLU
& $1.0\times10^{6}$ steps
& $\gamma=0.99$; replay buffer $=2\times10^{5}$; batch size $=144$; target update interval $=2000$; $\epsilon$: 1.0 to 0.05
& Penalty \\
RH-MILP
& Receding-horizon optimization& MILP binary variables
& Online re-solving
& $H_{\rm RH}=36$ steps; $\Delta t=60$ s; time limit $=0.25$ s; relative MIP gap $=0.01$; horizon discount $=0.995$
& Explicit \\
SA
& Metaheuristic
& Weight vector, 16-D
& 2 restarts; 80 iter./restart;
& Initial temperature $=2.5$; minimum temperature $=0.05$; cooling rate $=0.96$; proposal scale $=0.9$; 6 eval. eps./candidate
& Repair / penalty \\
GA
& Metaheuristic
& Weight chromosome, 16-D
& Population size  $=24$; 60 generations;& Elite size $=4$; tournament size $=3$; crossover rate $=0.85$; mutation rate $=0.20$; mutation scale $=0.60$; 6 eval. eps./individual
& Repair / penalty \\
\hline
\multicolumn{6}{@{}p{0.97\textwidth}@{}}{\footnotesize
All baselines follow the same evaluation protocol. ``eval. eps.'' denotes evaluation episodes. ``lr" denotes learning rate.}
\end{tabular}
\end{table*}

\begin{table*}[!t]
\centering
\caption{Quantitative comparison of different methods under varying target loads over a 10-day evaluation horizon. Both target completion rate and cumulative reward are reported as mean $\pm$ standard deviation over 20 independent Monte Carlo trials, and the 95\% confidence interval is reported for the mean cumulative reward.}
\label{tab:10day_results}
\renewcommand{\arraystretch}{1.12}
\setlength{\tabcolsep}{1.0pt}
\footnotesize
\begin{tabular}{c c c c c}
\hline
\makecell{Target load}& Method& \makecell{Target completion rate\\(mean $\pm$ std)} 
& \makecell{Cumulative reward\\(mean $\pm$ std)} 
& \makecell{95\% CI of reward}\\
\hline
20 & PCM-PPO & 0.705 $\pm$ 0.026 & 6217.5 $\pm$ 185.6 & [6130.6, 6304.4] \\
   & PB-PPO  & 0.593 $\pm$ 0.029 & 5428.4 $\pm$ 224.7 & [5323.2, 5533.6] \\
   & RH-MILP & 0.594 $\pm$ 0.027 & 5450.6 $\pm$ 206.8 & [5353.8, 5547.4] \\
   & PPO     & 0.586 $\pm$ 0.035 & 5365.9 $\pm$ 243.1 & [5252.1, 5479.7] \\
   & AC      & 0.569 $\pm$ 0.033 & 4927.2 $\pm$ 205.6 & [4831.0, 5023.4] \\
   & DQN     & 0.539 $\pm$ 0.032 & 4622.2 $\pm$ 211.3 & [4523.3, 4721.1] \\
   & SA      & 0.474 $\pm$ 0.036 & 4142.0 $\pm$ 192.4 & [4052.0, 4232.0] \\
   & GA      & 0.458 $\pm$ 0.038 & 3914.4 $\pm$ 218.7 & [3812.0, 4016.8] \\
\hline
30 & PCM-PPO & 0.667 $\pm$ 0.028& 8304.2 $\pm$ 238.5 & [8192.6, 8415.8] \\
   & PB-PPO  & 0.571 $\pm$ 0.033 & 7342.8 $\pm$ 286.9 & [7208.5, 7477.1] \\
   & RH-MILP & 0.563 $\pm$ 0.032 & 7228.6 $\pm$ 272.4 & [7101.1, 7356.1] \\
   & PPO     & 0.558 $\pm$ 0.038 & 7168.4 $\pm$ 318.6 & [7019.3, 7317.5] \\
   & AC      & 0.550 $\pm$ 0.036 & 7021.1 $\pm$ 259.7 & [6899.6, 7142.6] \\
   & DQN     & 0.526 $\pm$ 0.036 & 6713.3 $\pm$ 236.8 & [6602.5, 6824.1] \\
   & SA      & 0.468 $\pm$ 0.039 & 6208.0 $\pm$ 263.5 & [6084.7, 6331.3] \\
   & GA      & 0.451 $\pm$ 0.041 & 5963.7 $\pm$ 247.9 & [5847.7, 6079.7] \\
\hline
40 & PCM-PPO & 0.632 $\pm$ 0.031 & 11025.2 $\pm$ 326.4 & [10872.4, 11178.0] \\
   & PB-PPO  & 0.540 $\pm$ 0.037 & 9278.5 $\pm$ 371.6 & [9104.6, 9452.4] \\
   & RH-MILP & 0.520 $\pm$ 0.043 & 8958.4 $\pm$ 438.9 & [8753.0, 9163.8] \\
   & PPO     & 0.523 $\pm$ 0.043 & 8992.8 $\pm$ 425.3 & [8793.8, 9191.8] \\
   & AC      & 0.507 $\pm$ 0.041 & 8639.7 $\pm$ 315.8 & [8491.9, 8787.5] \\
   & DQN     & 0.486 $\pm$ 0.040 & 8100.6 $\pm$ 257.4 & [7980.1, 8221.1] \\
   & SA      & 0.432 $\pm$ 0.043 & 7507.7 $\pm$ 286.1 & [7373.8, 7641.6] \\
   & GA      & 0.419 $\pm$ 0.045 & 7330.9 $\pm$ 239.8 & [7218.7, 7443.1] \\
\hline
50 & PCM-PPO & 0.607 $\pm$ 0.034 & 13245.4 $\pm$ 421.8 & [13048.0, 13442.8] \\
   & PB-PPO  & 0.518 $\pm$ 0.040 & 11136.2 $\pm$ 501.7 & [10901.4, 11371.0] \\
   & RH-MILP & 0.493 $\pm$ 0.050 & 10690.2 $\pm$ 611.3 & [10404.1, 10976.3] \\
   & PPO     & 0.502 $\pm$ 0.047 & 10840.5 $\pm$ 552.1 & [10582.1, 11098.9] \\
   & AC      & 0.487 $\pm$ 0.044 & 10563.2 $\pm$ 411.6 & [10370.6, 10755.8] \\
   & DQN     & 0.474 $\pm$ 0.042 & 10055.7 $\pm$ 376.9 & [9879.3, 10232.1] \\
   & SA      & 0.430 $\pm$ 0.048 & 9326.8 $\pm$ 374.8 & [9151.4, 9502.2] \\
   & GA      & 0.419 $\pm$ 0.050 & 9099.6 $\pm$ 421.5 & [8902.3, 9296.9] \\
\hline
\end{tabular}
\end{table*}

To evaluate the effectiveness of PCM-PPO under different planning loads, comparative experiments were conducted against representative learning-based, optimization-based, and metaheuristic baselines, including PB-PPO, vanilla PPO, RH-MILP, AC, DQN, SA, and GA \cite{qi2024mission,wei2021deep,he2020generic,she2018onboard}. PB-PPO was included as a recent representative PPO-based mission-planning baseline, while PPO was used to provide a direct PPO comparison without the target-pool-based representation and physics-constrained masking design. RH-MILP was introduced as a receding-horizon constrained optimization baseline. AC and DQN were used as additional deep reinforcement-learning baselines, and SA and GA were included as representative metaheuristic scheduling baselines. The baseline configurations are summarized in \tabref{tab:baseline_settings}.

The settings in \tabref{tab:baseline_settings} specify the network and training configurations of the DRL baselines, the solver configuration of RH-MILP, and the heuristic parameters of SA and GA. For PB-PPO, the original actor--critic network with four hidden layers of 64 neurons was widened to four hidden layers of 256 neurons to avoid restricting its performance by insufficient network capacity in the present high-dimensional Titan mission-planning environment, while its main PPO-related hyperparameters were retained according to the reference configuration. The vanilla PPO, AC, and DQN baselines were trained with the same number of environment interaction steps as PCM-PPO. For RH-MILP, a finite-horizon MILP is constructed at each decision step using the current mission state and constraints on observation feasibility, downlink availability, resource evolution, and action exclusivity. The optimized sequence is implemented in a receding-horizon manner, where only the first action is executed before the horizon is shifted forward and the MILP is re-solved. For SA and GA, the heuristic encodings and algorithmic parameters listed in \tabref{tab:baseline_settings} were used to generate candidate schedules under the same simulation environment. The last column of \tabref{tab:baseline_settings} summarizes the corresponding constraint-handling mode, where penalty, explicit, and repair/penalty indicate reward-penalty feedback, direct constraint formulation, and heuristic repair with penalty treatment, respectively.

For reporting consistency, the cumulative reward denotes the total reward accumulated over a complete 10-day or 100-day evaluation horizon. The average daily reward is obtained by dividing the cumulative reward by the corresponding number of evaluation days and is used in the reward-comparison and box-plot figures. Except for the sensitivity metric explicitly identified as nominally rescored in \tabref{tab:reward_sensitivity}, all reported rewards are evaluated using the reward coefficients of the corresponding policy setting.

\begin{table}[t]
\centering
\caption{Opportunity-target completion rate of representative methods over the 10-day evaluation horizon.}
\label{tab:opp_completion_10d}
\scriptsize
\setlength{\tabcolsep}{3pt}
\renewcommand{\arraystretch}{1.08}
\resizebox{\columnwidth}{!}{
\begin{tabular}{lcccc}
\hline
Algorithm & Targets 20 & Targets 30 & Targets 40 & Targets 50 \\
\hline
PCM-PPO & 0.802 & 0.758 & 0.704 & 0.660 \\
PB-PPO  & 0.764 & 0.717 & 0.668 & 0.608 \\
RH-MILP & 0.769 & 0.701 & 0.645 & 0.589 \\
PPO     & 0.718 & 0.682 & 0.642 & 0.593 \\
AC      & 0.693 & 0.655 & 0.614 & 0.567 \\
\hline
\end{tabular}}
\vspace{1mm}

\parbox{\columnwidth}{\footnotesize
The column headers denote daily target loads. The rate is computed with respect to generated opportunity targets only.}
\end{table}

\begin{figure}[!t]
  \centering
  \includegraphics[width=7.5cm]{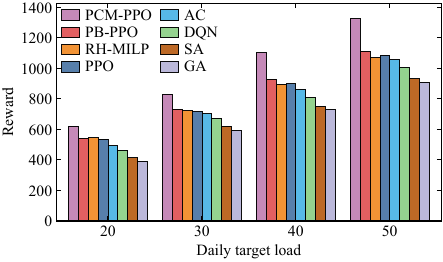}
  \caption{Average daily reward comparison under different daily target loads over the 10-day evaluation horizon.}
  \label{fig_4_3}
\end{figure}

\begin{figure*}[!t]
  \centering
  \includegraphics[width=16.0cm]{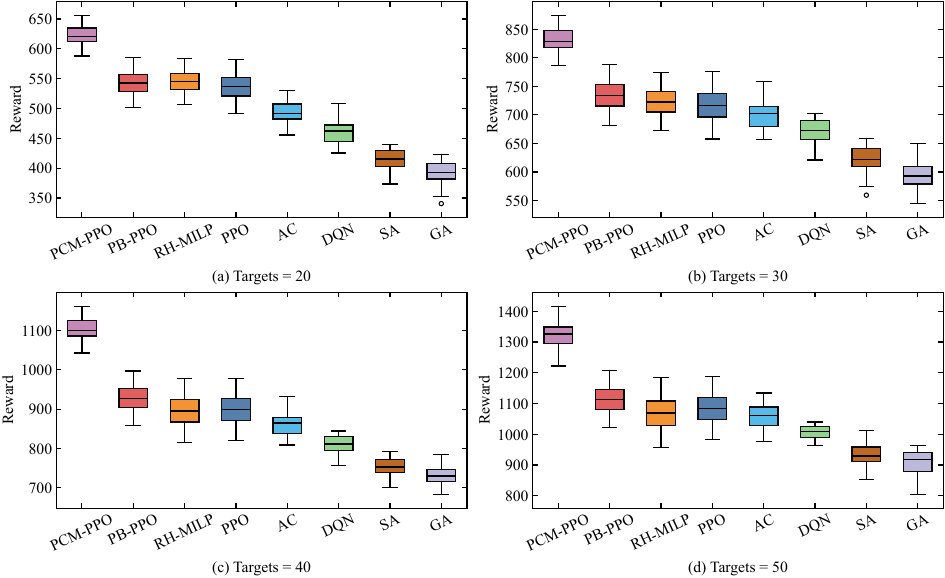}
  \caption{Box plots of average daily rewards over 20 independent Monte Carlo trials under the 10-day evaluation horizon.}
  \label{fig_4_4}
\end{figure*}

The 10-day evaluation results are summarized in \tabref{tab:10day_results}, \figref{fig_4_3}, \figref{fig_4_4}, and \tabref{tab:opp_completion_10d}. In \tabref{tab:10day_results}, the target completion rate and cumulative reward are reported as mean $\pm$ standard deviation over 20 independent Monte Carlo trials, and the 95\% confidence interval is provided for the mean cumulative reward. \figref{fig_4_3} presents the corresponding average daily reward comparison under different daily target loads, while \figref{fig_4_4} further presents the average daily reward distributions over the 20 Monte Carlo trials. 
To further examine the response capability to dynamically arriving time-sensitive targets, \tabref{tab:opp_completion_10d} reports the opportunity-target completion rate for representative methods. This metric is computed with respect to generated opportunity targets only.

As shown in \tabref{tab:10day_results} and \figref{fig_4_3}, PCM-PPO achieves the highest cumulative reward and target completion rate under all daily target-load settings. Its advantage is maintained as the daily target load increases from 20 to 50, indicating that the proposed method remains effective when the number of competing observation opportunities and resource-allocation conflicts increases. For instance, under the daily target load of 50, PCM-PPO obtains a cumulative reward of $13245.4\pm421.8$, corresponding to an average daily reward of 1324.5, and its target completion rate reaches $0.607\pm0.034$. In comparison, PB-PPO, PPO, and RH-MILP achieve cumulative rewards of $11136.2\pm501.7$, $10840.5\pm552.1$, and $10690.2\pm611.3$, respectively, with lower completion rates of $0.518\pm0.040$, $0.502\pm0.047$, and $0.493\pm0.050$. Moreover, \tabref{tab:opp_completion_10d} shows that PCM-PPO also achieves the highest opportunity-target completion rate among the representative methods. Under the daily target load of 50, PCM-PPO achieves an opportunity-target completion rate of 0.660, remaining above PB-PPO, RH-MILP, PPO, and AC, which obtain 0.608, 0.589, 0.593, and 0.567, respectively. These results show that the performance gain of PCM-PPO is not only reflected in overall reward accumulation, but also in its ability to respond to dynamically arriving time-sensitive targets.

The performance differences among the compared methods can be further explained from the perspective of dynamic task handling and constraint-aware decision making. PB-PPO achieves competitive performance among the baselines, indicating that PPO-based policy optimization is effective for sequential mission planning. However, PB-PPO still lacks an explicit target-conditioned mechanism to evaluate dynamically changing target slots, and therefore its response to newly arrived opportunity targets is limited under coupled constraints. PPO shows a more pronounced limitation, because infeasible or low-return actions are mainly discouraged through reward penalties rather than being excluded before action selection. Consequently, PPO-based methods become less effective than PCM-PPO when the target load increases and the feasible action set changes frequently. RH-MILP explicitly enforces the observation, downlink, resource, and action-exclusivity constraints within each receding horizon, and therefore performs competitively under relatively low target loads. Nevertheless, its performance is affected by finite-horizon approximation and repeated re-optimization. As the target load increases, more candidate targets and opportunity-target arrivals must be considered within a limited horizon, making RH-MILP more sensitive to stochastic visibility, utility decay, and target-arrival timing. For AC, the strongly constrained action space leads to high-variance policy-gradient updates, especially when many sampled actions are infeasible or low-return under the current mission state. For DQN, the value network must estimate action values over a large number of time-varying feasible and infeasible discrete actions, which makes multi-step value estimation more difficult under dynamic feasibility conditions. For SA and GA, the candidate scheduling space expands rapidly with the number of targets, and the repair/penalty mechanism becomes less effective in consistently identifying high-quality feasible schedules under coupled observation, downlink, and resource constraints. Overall, PCM-PPO gains its advantage by using a bounded dynamic target representation to accommodate opportunity targets, target-conditioned policy inference to evaluate current target priorities, and physics-constrained feasible action filtering to restrict decisions to executable actions in the Titan mission planning environment.

In addition to the mean performance reported in \tabref{tab:10day_results} and \figref{fig_4_3}, \figref{fig_4_4} further illustrates the average daily reward distributions over 20 independent Monte Carlo trials. PCM-PPO achieves the highest median average daily reward under all target-load settings, indicating that its advantage is not caused by a few favorable scenarios. Under higher target loads, especially 40 and 50, the separation between PCM-PPO and several baselines becomes more evident, while some baselines exhibit wider reward distributions. Denser target competition and stronger coupled constraints make planning less stable for the competing methods. Nevertheless, PCM-PPO maintains a higher and relatively concentrated reward distribution compared with the other methods. These results further confirm that PCM-PPO provides more reliable 10-day planning performance under Titan mission scenarios.

\begin{table*}[!t]
\centering
\caption{Quantitative comparison of different methods under varying target loads over a 100-day evaluation horizon. Both target completion rate and cumulative reward are reported as mean $\pm$ standard deviation over 20 independent Monte Carlo trials, and the 95\% confidence interval is reported for the mean cumulative reward.}
\label{tab:100day_results}
\renewcommand{\arraystretch}{1.12}
\setlength{\tabcolsep}{3.0pt}
\footnotesize
\begin{tabular}{c c c c c}
\hline
\makecell{Target load}& Algorithm 
& \makecell{Target completion rate\\(mean $\pm$ std)} 
& \makecell{Cumulative reward\\(mean $\pm$ std)} 
& \makecell{95\% CI of reward}\\
\hline
20 & PCM-PPO & 0.671 $\pm$ 0.011 & 58480.6 $\pm$ 965.4 & [58028.8, 58932.4] \\
   & PB-PPO  & 0.584 $\pm$ 0.013 & 51986.7 $\pm$ 1056.4 & [51492.3, 52481.1] \\
   & RH-MILP & 0.575 $\pm$ 0.012 & 51320.8 $\pm$ 1034.6 & [50836.6, 51805.0] \\
   & PPO     & 0.572 $\pm$ 0.014 & 50642.3 $\pm$ 1158.9 & [50099.9, 51184.7] \\
   & AC      & 0.564 $\pm$ 0.013 & 47170.9 $\pm$ 1048.5 & [46680.2, 47661.6] \\
   & DQN     & 0.532 $\pm$ 0.013 & 45982.1 $\pm$ 1015.6 & [45506.8, 46457.4] \\
   & SA      & 0.459 $\pm$ 0.015 & 39724.5 $\pm$ 1226.4 & [39150.5, 40298.5] \\
   & GA      & 0.443 $\pm$ 0.016 & 38216.4 $\pm$ 1308.6 & [37604.0, 38828.8] \\
\hline
30 & PCM-PPO & 0.621 $\pm$ 0.013& 81365.2 $\pm$ 1376.8 & [80720.8, 82009.6] \\
   & PB-PPO  & 0.560 $\pm$ 0.015 & 74685.4 $\pm$ 1568.2 & [73951.5, 75419.3] \\
   & RH-MILP & 0.542 $\pm$ 0.016 & 71610.5 $\pm$ 1668.9 & [70829.4, 72391.6] \\
   & PPO     & 0.543 $\pm$ 0.017 & 71642.8 $\pm$ 1764.3 & [70817.1, 72468.5] \\
   & AC      & 0.535 $\pm$ 0.016 & 68550.3 $\pm$ 1542.6 & [67828.3, 69272.3] \\
   & DQN     & 0.510 $\pm$ 0.016 & 65237.1 $\pm$ 1498.3 & [64535.9, 65938.3] \\
   & SA      & 0.454 $\pm$ 0.019 & 59166.8 $\pm$ 1806.5 & [58321.3, 60012.3] \\
   & GA      & 0.425 $\pm$ 0.020 & 55842.3 $\pm$ 1904.2 & [54951.1, 56733.5] \\
\hline
40 & PCM-PPO & 0.596 $\pm$ 0.016 & 99580.5 $\pm$ 2018.5 & [98635.8, 100525.2] \\
   & PB-PPO  & 0.529 $\pm$ 0.018 & 91386.5 $\pm$ 2314.6 & [90303.2, 92469.8] \\
   & RH-MILP & 0.498 $\pm$ 0.023 & 84850.6 $\pm$ 2864.5 & [83510.0, 86191.2] \\
   & PPO     & 0.506 $\pm$ 0.021 & 86642.4 $\pm$ 2678.2 & [85389.0, 87895.8] \\
   & AC      & 0.494 $\pm$ 0.020 & 81698.6 $\pm$ 2316.7 & [80614.4, 82782.8] \\
   & DQN     & 0.482 $\pm$ 0.020 & 78945.9 $\pm$ 2218.4 & [77907.7, 79984.1] \\
   & SA      & 0.405 $\pm$ 0.024 & 67254.6 $\pm$ 2685.9 & [65997.6, 68511.6] \\
   & GA      & 0.391 $\pm$ 0.025 & 64857.2 $\pm$ 2856.3 & [63520.4, 66194.0] \\
\hline
50 & PCM-PPO & 0.575 $\pm$ 0.018 & 120560.8 $\pm$ 2864.2 & [119220.3, 121901.3] \\
   & PB-PPO  & 0.509 $\pm$ 0.022 & 110842.6 $\pm$ 3316.8 & [109290.3, 112394.9] \\
   & RH-MILP & 0.473 $\pm$ 0.029 & 98680.2 $\pm$ 4382.7 & [96629.1, 100731.3] \\
   & PPO     & 0.486 $\pm$ 0.025 & 104736.9 $\pm$ 3896.4 & [102913.3, 106560.5] \\
   & AC      & 0.478 $\pm$ 0.024 & 99613.8 $\pm$ 3418.9 & [98013.7, 101213.9] \\
   & DQN     & 0.471 $\pm$ 0.023 & 98072.7 $\pm$ 3296.5 & [96529.9, 99615.5] \\
   & SA      & 0.393 $\pm$ 0.028 & 81564.9 $\pm$ 3862.5 & [79757.2, 83372.6] \\
   & GA      & 0.385 $\pm$ 0.030 & 80116.8 $\pm$ 4088.7 & [78203.3, 82030.3] \\
\hline
\end{tabular}
\end{table*}

\begin{figure}[!t]
  \centering
  \includegraphics[width=7.5cm]{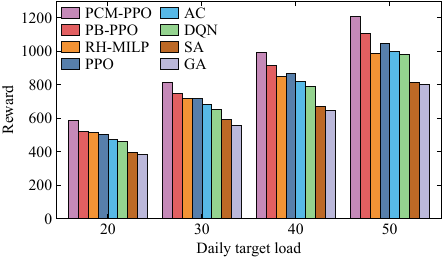}
  \caption{Average daily reward comparison under different daily target loads over the 100-day evaluation horizon.}
  \label{fig_4_5}
\end{figure}
\begin{figure*}[!t]
  \centering
  \includegraphics[width=16.0cm]{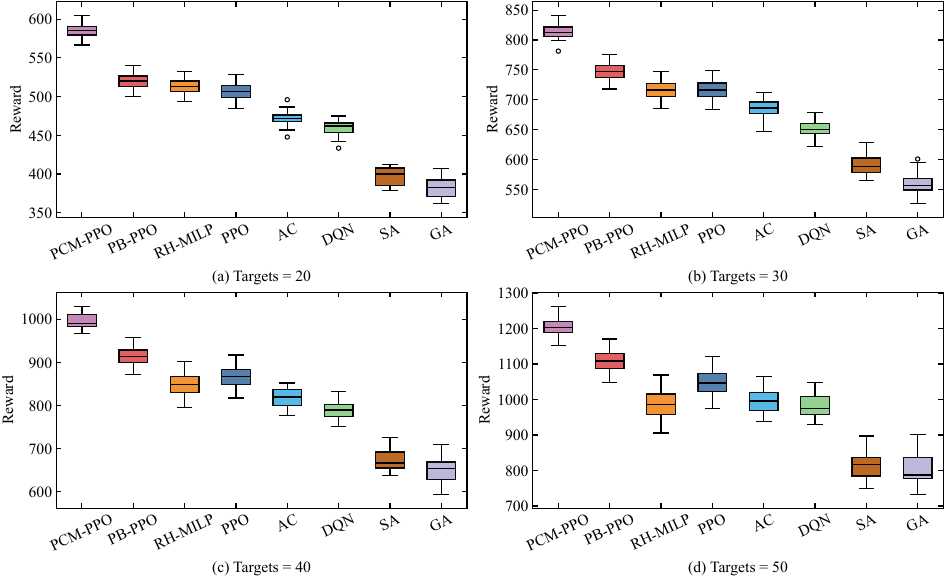}
  \caption{Box plots of average daily rewards over 20 independent Monte Carlo trials under the 100-day evaluation horizon.}
  \label{fig_4_6}
\end{figure*}

To further evaluate the capability of the proposed method for long-horizon autonomous mission planning, the evaluation horizon was extended to 100 days. The quantitative results are reported in \tabref{tab:100day_results}, and the corresponding average daily reward comparison under different daily target loads is shown in \figref{fig_4_5}. In addition, \figref{fig_4_6} presents the average daily reward distributions over 20 independent Monte Carlo trials, and \tabref{tab:training_seed_stability} further reports the training-seed stability of PPO-based methods under the representative target load of 30.

As shown in \tabref{tab:100day_results} and \figref{fig_4_5}, PCM-PPO continues to achieve the highest cumulative reward and target completion rate under all daily target-load settings over the 100-day evaluation horizon. For example, when the daily target load is 50, PCM-PPO obtains a cumulative reward of $120560.8\pm2864.2$ and a target completion rate of $0.575\pm0.018$, outperforming PB-PPO, RH-MILP, PPO, AC, DQN, SA, and GA. Compared with the 10-day evaluation, all methods exhibit a certain degree of performance degradation as the evaluation horizon becomes longer, which is reasonable because stochastic cloud visibility, DSN-window variation, utility decay, and resource evolution accumulate over more decision episodes. Nevertheless, PCM-PPO maintains a clear advantage, indicating that the proposed method can preserve effective planning performance under extended autonomous operation.

The 100-day evaluation further reflects how different planning formulations respond to accumulated stochastic mission variations. PB-PPO and PPO can learn useful scheduling policies, but repeated changes in target availability, utility states, and onboard resources make their penalty-based decisions less stable over the extended horizon. RH-MILP remains feasible at each decision step because the constraints are explicitly enforced, yet its finite-horizon approximation and repeated re-optimization make it difficult to maintain consistent decisions across successive stochastic daily scenarios. For AC and DQN, the accumulation of stochastic transitions further increases the difficulty of policy-gradient optimization and value estimation under time-varying feasible-action patterns. The metaheuristic baselines are also affected by the repeated expansion of the candidate scheduling space under long-horizon planning. PCM-PPO performs better in this long-horizon setting because its policy repeatedly evaluates the current target state and executable action domain at each decision step rather than relying only on penalty feedback or finite-horizon re-optimization.

The reward distributions in \figref{fig_4_6} further support this observation. PCM-PPO achieves the highest median average daily reward under all target-load settings, and its distribution remains above those of the competing methods. Under higher target loads, especially 40 and 50, the separation between PCM-PPO and several baselines becomes more evident, while some baselines exhibit wider reward distributions. This indicates that denser target competition and accumulated stochastic variations make long-horizon planning less stable for methods that rely mainly on penalty feedback, finite-horizon re-optimization, or heuristic repair. In contrast, PCM-PPO maintains a relatively high and concentrated reward distribution, suggesting more reliable long-horizon planning behavior across independent Monte Carlo trials.

To further quantify the practical magnitude of the main performance differences, standardized mean differences were computed between PCM-PPO and PB-PPO, which is the closest PPO-based baseline in the comparative experiments. The standardized effect size was calculated as the mean cumulative-reward difference divided by the pooled standard deviation of the two methods. Across daily target loads from 20 to 50, the standardized effect sizes range from 3.64 to 4.99 over the 10-day evaluation horizon and from 3.14 to 6.42 over the 100-day evaluation horizon. The corresponding mean reward improvements range from 13.1\% to 18.9\% and from 8.8\% to 12.5\%, respectively. These results indicate that the performance gains of PCM-PPO are not only consistent in terms of mean reward and confidence intervals, but also practically substantial in magnitude.

Finally, \tabref{tab:training_seed_stability} reports the training-seed stability of the PPO-based methods under the representative target load of 30. Five independently trained policies were obtained for each method, and each trained policy was evaluated on the same 20 Monte Carlo mission scenarios. PCM-PPO achieves the highest mean cumulative reward and target completion rate, with $81272.6\pm686.3$ and $0.619\pm0.009$, respectively. In comparison, PB-PPO and PPO obtain lower mean rewards and larger training-seed variations, with reward standard deviations of 842.7 and 976.5 and completion-rate standard deviations of 0.010 and 0.012, respectively. These results indicate that PCM-PPO is less sensitive to training randomness in the tested long-horizon setting, supporting the stability of the reported PPO-based comparison.

In summary, these results confirm that PCM-PPO is effective for Titan autonomous mission planning and remains reliable as the target load and evaluation horizon increase.

\begin{table}[!t]
\centering
\caption{Training-seed stability of PPO-based methods under target load 30 over the 100-day evaluation horizon.}
\label{tab:training_seed_stability}
\renewcommand{\arraystretch}{1.12}
\setlength{\tabcolsep}{4pt}
\footnotesize
\begin{tabular}{c c c c}
\hline
Method& \makecell{Training\\seeds} 
& \makecell{Cumulative reward\\mean $\pm$ std} 
& \makecell{Target completion rate\\mean $\pm$ std} \\
\hline
PCM-PPO & 5 & 81272.6 $\pm$ 686.3& 0.619 $\pm$ 0.009\\
PB-PPO  & 5 & 74591.8 $\pm$ 842.7& 0.558 $\pm$ 0.010\\
PPO     & 5 & 71584.4 $\pm$ 976.5& 0.541 $\pm$ 0.012\\
\hline
\end{tabular}
\end{table}

\subsection{Ablation study and sensitivity analysis}\label{4.4}

\subsubsection{Ablation study}\label{4.4.1}

\begin{figure}[!t]
  \centering
  \includegraphics[width=7.5cm]{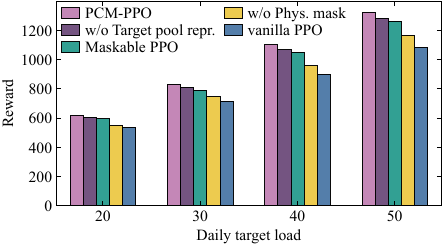}
  \caption{Average daily reward comparison under different ablation settings over the 10-day evaluation horizon.}
  \label{fig4_7}
\end{figure}
\begin{figure*}[!t]
  \centering
  \includegraphics[width=16.0cm]{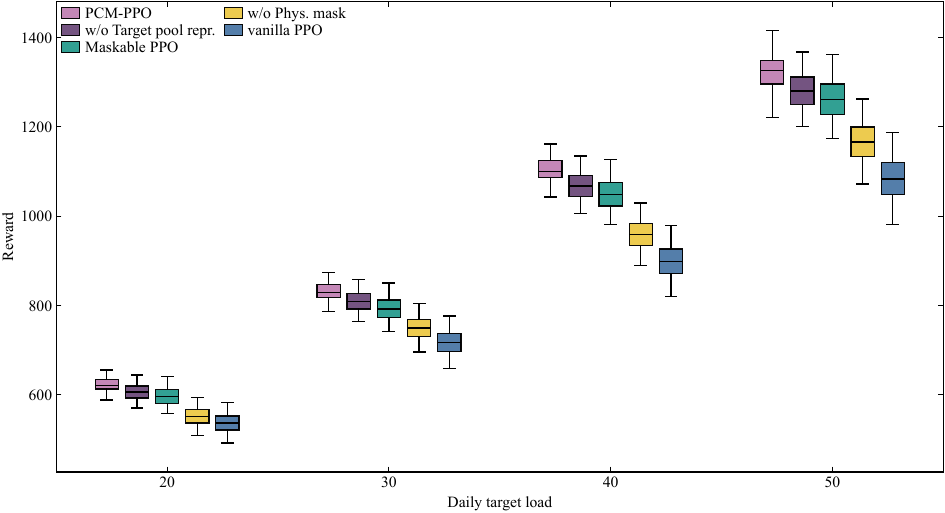}
  \caption{Box plots of average daily rewards under different ablation settings over the 10-day evaluation horizon.}
  \label{fig4_8}
\end{figure*}

To further investigate the contribution of the key components in PCM-PPO, an ablation study was conducted by comparing the full PCM-PPO method with four representative ablation and reference variants, namely w/o Target pool repr., Maskable PPO, w/o Phys. mask, and vanilla PPO. The w/o Target pool repr. variant replaces the fixed-capacity target pool with a step-wise padded candidate representation, in which currently available candidate targets are encoded directly at each decision step rather than represented through occupied or empty target slots. Besides, it uses a conventional flattened MLP actor while retaining the physics-constrained action mask and the integrated training strategy. Maskable PPO denotes a standard maskable PPO implementation that uses the same physics-constrained action mask but adopts a conventional flattened MLP actor. The w/o Phys. mask variant retains the target-pool representation, target-conditioned actor, and integrated training strategy, but removes the physics-constrained action mask and handles infeasible decisions mainly through reward-penalty feedback. Vanilla PPO denotes the original PPO algorithm, without the physics-constrained action mask, target-pool-based structured representation, and training strategy, used as the fully unmasked PPO reference. All variants were evaluated under the same Titan mission-planning scenarios, target-load settings, and Monte Carlo protocol described in \secref{4.3}.

The ablation results over the 10-day evaluation horizon are shown in \figref{fig4_7} and \figref{fig4_8}. As shown in \figref{fig4_7}, PCM-PPO achieves the highest average daily reward under all daily target-load settings, indicating that the full combination of fixed-capacity target-pool representation, target-conditioned policy inference, physics-constrained action masking, and integrated training strategy provides the best planning performance. The w/o Target pool repr. variant obtains the second-best performance because it still retains the physics-constrained action mask and the integrated training strategy. However, its reward remains consistently lower than that of PCM-PPO, especially under higher target loads, suggesting that organizing predefined and opportunity targets through a fixed-capacity target pool and representing the mission state with a target-wise state matrix are beneficial for target handling. Maskable PPO performs worse than w/o Target pool repr., which indicates that the physics-constrained action mask with a conventional flattened actor is not sufficient to recover the benefit of the target-pool-based state representation and target-conditioned actor. The w/o Phys. mask variant suffers a more evident performance reduction, showing that reward-penalty feedback is less effective than explicit feasibility filtering in the strongly constrained Titan mission-planning environment. Vanilla PPO produces the lowest reward among the compared variants, further indicating that removing the target-pool-based representation, target-conditioned policy modeling, physics-constrained action masking, and integrated training strategy simultaneously leads to the most severe performance degradation, confirming the effectiveness of key components in PCM-PPO.

The reward distributions in \figref{fig4_8} further support the above observations. PCM-PPO maintains the highest median average daily reward under all target-load settings, indicating that its advantage is not caused by a few favorable Monte Carlo scenarios. As the daily target load increases, the reward distributions of the ablated variants become more dispersed, especially for w/o Phys. mask and vanilla PPO. This suggests that removing the feasibility filtering mechanism makes the planning results more sensitive to stochastic target visibility, resource evolution, and dynamically changing feasible-action sets. Although w/o Target pool repr. and Maskable PPO remain more stable than the fully unmasked variants, their median rewards are still consistently lower than that of PCM-PPO. These results further indicate that the proposed key components improve not only the average planning reward but also the consistency of mission planning performance across stochastic Monte Carlo scenarios.

\begin{figure}[!t]
  \centering
  \includegraphics[width=7.5cm]{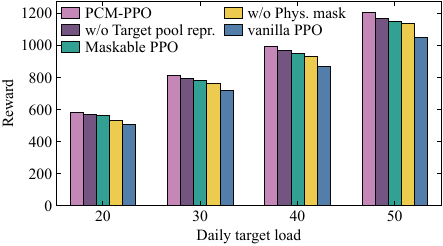}
  \caption{Average daily reward comparison under different ablation settings over the 100-day evaluation horizon.}
  \label{fig4_9}
\end{figure}
\begin{figure*}[!t]
  \centering
  \includegraphics[width=16.0cm]{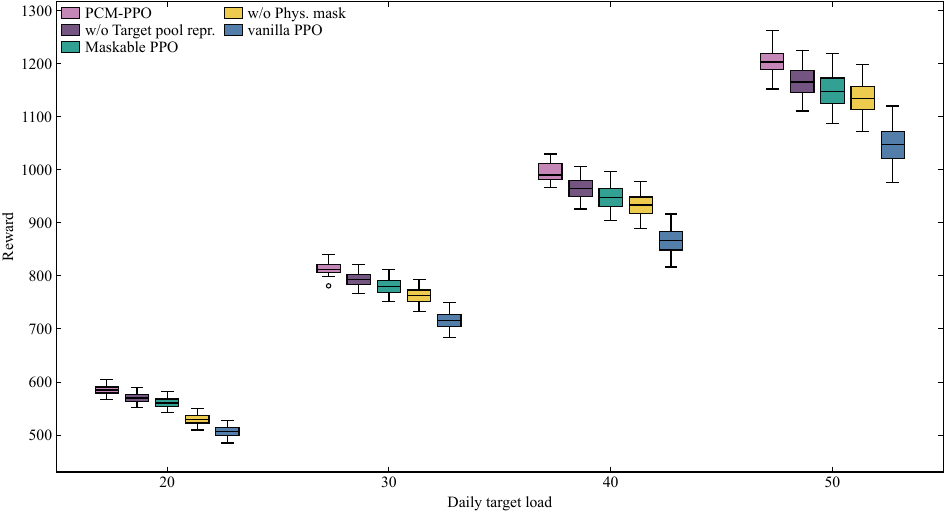}
  \caption{Box plots of average daily rewards under different ablation settings over the 100-day evaluation horizon.}
  \label{fig4_10}
\end{figure*}
To further examine whether the contribution of each component can be preserved in long-horizon autonomous mission planning, the ablation study was extended to the 100-day evaluation horizon. The corresponding results are shown in \figref{fig4_9} and \figref{fig4_10}. As shown in \figref{fig4_9}, PCM-PPO still achieves the highest average daily reward under all daily target-load settings, while vanilla PPO produces the lowest reward among the compared variants. The performance order of the ablation variants remains generally consistent with that observed in the 10-day evaluation, indicating that the effects of the proposed components are not limited to short evaluation horizons. In particular, w/o Target pool repr. remains below PCM-PPO, suggesting that the fixed-capacity target-pool representation and target-wise state matrix continue to benefit target organization under extended mission planning. Maskable PPO performs lower than w/o Target pool repr., suggesting that it is less effective for maintaining target-level prioritization as target availability, utility, and onboard resource states evolve over a longer planning horizon. The larger performance gap of w/o Phys. mask further indicates that explicit feasibility filtering becomes increasingly important when infeasible or low-return decisions may accumulate over long-horizon mission planning.

The reward distributions in \figref{fig4_10} further demonstrate the stability of the ablation results over the 100-day evaluation horizon. PCM-PPO maintains the highest median average daily reward under all target-load settings, and its reward distribution remains above those of the ablated variants. As the daily target load increases, the reward distributions of w/o Phys. mask and vanilla PPO become more dispersed, indicating that the absence of explicit feasibility filtering makes planning performance more sensitive to accumulated stochastic visibility changes, resource evolution, and time-varying feasible-action sets. In comparison, w/o Target pool repr. and Maskable PPO exhibit more stable distributions than the fully unmasked variants, but their median rewards remain consistently lower than that of PCM-PPO. These results further show that the proposed core components in PCM-PPO improve not only average planning reward, but also the consistency of mission-planning performance under long-horizon stochastic scenarios.

\begin{figure}[!t]
  \centering
  \includegraphics[width=7.5cm]{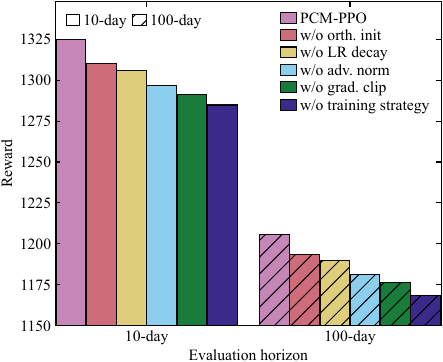}
  \caption{Average daily reward comparison under different training strategy settings at daily target load 50.}
  \label{fig4_11}
\end{figure}

In addition to the structural ablation settings, the contribution of the integrated training strategy was further examined at the daily target load of 50, as shown in \figref{fig4_11}. The full PCM-PPO method achieves the highest average daily reward under both the 10-day and 100-day evaluation horizons. Removing each training component leads to a certain reward reduction, indicating that orthogonal initialization, learning-rate decay, advantage normalization, and gradient clipping all contribute to policy optimization under the high-dimensional mission state and strongly constrained action space. Among these variants, w/o Training strategy produces the lowest reward, showing that removing all training-stabilization components simultaneously has the most evident negative effect. Nevertheless, the performance gaps among the single-component removal settings are smaller than those caused by removing the target-pool representation or the physics-constrained mask, which suggests that the integrated training strategy mainly improves optimization stability and reward acquisition rather than changing the fundamental feasible-action structure.

Overall, the ablation results consistently verify the effectiveness of the key components in PCM-PPO across different target loads and evaluation horizons, and demonstrate their combined contribution to planning performance.

\subsubsection{Sensitivity analysis}\label{4.4.2}

\begin{table*}[t]
\centering
\caption{Planning-quality sensitivity of PCM-PPO to reward-coefficient perturbations at daily target load 50 over the 10-day evaluation horizon.}
\label{tab:reward_sensitivity}
\small
\setlength{\tabcolsep}{4pt}
\renewcommand{\arraystretch}{1.10}
\begin{tabular}{lccccc}
\hline
Reward setting
& Nominal avg. reward& Target completion rate
& Downlink ratio
& Avg. peak storage occupancy& Min. energy margin (Wh) \\
\hline
Nominal & 1324.5 & 0.607 & 0.821 & 0.86 & 246 \\
$\lambda_{\mathrm{obs}}$ $-20\%$ & 1289.8 & 0.590 & 0.836 & 0.81 & 269\\
$\lambda_{\mathrm{obs}}$ $+20\%$ & 1336.2 & 0.617 & 0.803 & 0.92 & 224 \\
$\lambda_{\mathrm{dl}}$ $-20\%$ & 1328.7& 0.611 & 0.769 & 0.93& 235 \\
$\lambda_{\mathrm{dl}}$ $+20\%$ & 1316.8 & 0.598 & 0.861 & 0.77& 244\\
$(\lambda_E,\lambda_S)$ $-20\%$& 1331.4 & 0.613 & 0.814 & 0.93 & 210\\
$(\lambda_E,\lambda_S)$ $+20\%$& 1297.6 & 0.595 & 0.829 & 0.80 & 304 \\
$\lambda_{\mathrm{idle}}$ $-20\%$ & 1322.9 & 0.606 & 0.820 & 0.87 & 247 \\
$\lambda_{\mathrm{idle}}$ $+20\%$ & 1321.7 & 0.608 & 0.817 & 0.86 & 240 \\
\hline
\multicolumn{6}{p{0.94\textwidth}}{\footnotesize
The average peak storage occupancy is normalized by the onboard storage capacity. Nominal avg. reward denotes nominally rescored average daily reward.}
\end{tabular}
\end{table*}

To further examine the sensitivity of PCM-PPO to reward-coefficient settings and feasibility filtering, additional evaluations were conducted at the daily target load of 50 over the 10-day evaluation horizon. For each reward setting, PCM-PPO was trained using the same training random seed and evaluated on the same 20 independent Monte Carlo mission scenarios. Thus, all values in \tabref{tab:reward_sensitivity} are averages over the Monte Carlo scenarios, but are not additionally averaged over training seeds. Training-seed variability is examined separately in \tabref{tab:training_seed_stability}. The reward-coefficient sensitivity results are summarized in \tabref{tab:reward_sensitivity}. For each perturbed setting, one reward component was scaled by $\pm20\%$ while the other coefficients were kept unchanged. Since different reward coefficients change the numerical reward scale, the reported nominally rescored average daily reward was recomputed using the nominal reward coefficients for all evaluated policies. In addition to the average daily reward and target completion rate, the downlink ratio, average peak storage occupancy, and minimum energy margin were reported to provide a more detailed evaluation of planning quality and resource behavior.

As shown in \tabref{tab:reward_sensitivity}, the perturbation of $\lambda_{\mathrm{obs}}$ directly affects the trade-off between scientific observation return and resource preservation. Increasing $\lambda_{\mathrm{obs}}$ by 20\% encourages a more aggressive observation policy. However, compared with the nominal setting, the nominally rescored average daily reward only increases from 1324.5 to 1336.2, and the target completion rate only increases from 0.607 to 0.617. Meanwhile, the average peak storage occupancy rises to 0.92, approaching the onboard storage limit, and the minimum energy margin decreases to 224 Wh, which is closer to the energy safety margin. This indicates that further emphasizing observation reward yields only limited additional planning gain while noticeably increasing resource pressure. In contrast, decreasing $\lambda_{\mathrm{obs}}$ by 20\% leads to a clear reduction in the nominally rescored average daily reward and target completion rate, which decrease to 1289.8 and 0.590, respectively, but it preserves larger resource margins. Therefore, the nominal observation-reward coefficient provides a more balanced setting: it avoids overly conservative observation behavior while also preventing an aggressive observation policy from approaching resource safety limits for only marginal reward improvement.

As shown in \tabref{tab:reward_sensitivity}, reducing $\lambda_{\mathrm{dl}}$ by 20\% slightly increases the target completion rate from 0.607 to 0.611, since fewer decision opportunities are encouraged to be assigned to downlink actions. However, the nominally rescored average daily reward only reaches 1328.7, and the downlink ratio decreases evidently from 0.821 to 0.769. Meanwhile, the average peak storage occupancy increases to 0.93, showing that insufficient emphasis on downlink actions weakens data return and increases onboard storage pressure. Conversely, increasing $\lambda_{\mathrm{dl}}$ by 20\% improves the downlink ratio to 0.861 and reduces the average peak storage occupancy to 0.77, but the nominally rescored average daily reward decreases to 1316.8 and the target completion rate drops to 0.598. The result indicates that an overly large downlink reward may over-prioritize data transmission and reduce the opportunity for valuable observations. Therefore, the nominal downlink-reward coefficient provides a more balanced setting by maintaining effective data return and storage relief without unnecessarily sacrificing observation performance.

The perturbation of the resource-penalty coefficients further shows the balance between observation aggressiveness and resource preservation. Reducing $(\lambda_E,\lambda_S)$ by 20\% increases the nominally rescored average daily reward to 1331.4 and the target completion rate to 0.613, but this improvement is obtained with a higher average peak storage occupancy of 0.93 and a reduced minimum energy margin of 210 Wh, which is very close to the prescribed energy safety margin. In contrast, increasing $(\lambda_E,\lambda_S)$ by 20\% makes the policy more conservative: the minimum energy margin increases to 304 Wh and the average peak storage occupancy decreases to 0.80, but the nominally rescored average daily reward and target completion rate drop to 1297.6 and 0.595, respectively. These results indicate that the nominal resource-penalty coefficients avoid both unsafely excessive resource consumption and overly conservative planning. In addition, perturbing $\lambda_{\mathrm{idle}}$ by $\pm20\%$ causes only minor changes in all reported metrics, indicating that the idle penalty mainly discourages unnecessary standby actions without dominating the planning objective.

\begin{table}[t]
\centering
\caption{Infeasible-action ratios relative to the modeled admissible action set before and after masking at a daily target load of 50 over the 10-day evaluation horizon.}
\label{tab:infeasible_action_ratio}
\small
\setlength{\tabcolsep}{4pt}
\renewcommand{\arraystretch}{1.12}
\begin{tabular}{lcc}
\hline
Method variant& \makecell{Raw-policy\\infeasible ratio}& \makecell{Final selected\\infeasible ratio}\\
\hline
PCM-PPO & 0.107& 0.000 \\
Maskable PPO & 0.123& 0.000 \\
w/o Phys. mask & 0.052& 0.052\\
vanilla PPO & 0.078& 0.078\\
\hline
\end{tabular}
\end{table}

Moreover, \tabref{tab:infeasible_action_ratio} reports an action-mask enforcement check at a daily target load of 50 over the 10-day evaluation horizon. The raw-policy infeasible ratio measures the proportion of actions preferred by the unfiltered policy output that do not belong to the modeled admissible action set. The final selected infeasible ratio measures the proportion of selected actions that remain outside this same admissible set after masking or final action selection.

For PCM-PPO and Maskable PPO, the final selected infeasible ratio is reduced to $0.000$. This result is an expected consequence of deterministic hard masking: actions with zero mask entries are assigned zero probability and cannot be sampled by the masked policy. Therefore, the zero ratio verifies that the implemented masking procedure enforces action selection within the modeled admissible set. The nonzero raw-policy infeasible ratios indicate that, without this filtering, the policy output would prefer inadmissible actions at a non-negligible fraction of decision steps. In contrast, w/o Phys.\ mask and vanilla PPO retain nonzero final selected infeasible ratios of $0.052$ and $0.078$, respectively, because their selected actions are not restricted by the explicit physical feasibility filter.

Accordingly, \tabref{tab:infeasible_action_ratio} should be interpreted as an implementation-level feasibility-enforcement check rather than independent empirical evidence of operational safety. Because PCM-PPO and Maskable PPO both use deterministic hard masking, their identical zero final ratios are expected and do not establish a comparative operational-safety advantage of PCM-PPO. The table also does not account for model mismatch, state-estimation errors, actuation faults, or other flight-software and hardware failures.

To evaluate the sensitivity of the RH-MILP baseline to its online planning configuration, we conduct a configuration sensitivity analysis at a daily target load of 50 over the 10-day evaluation horizon. The nominal RH-MILP configuration uses a receding-horizon length of $H_{\mathrm{RH}}=36$ decision steps, a solver time limit of $0.25$ s per decision step, and a relative MIP gap of $1\%$. The sensitivity of the receding-horizon length, solver time limit, and relative MIP gap is assessed separately, while the remaining RH-MILP settings are retained. Each entry is averaged over five independent Monte Carlo trials.

\begin{table}[!t]
\centering
\caption{RH-MILP configuration sensitivity at a daily target load of 50.}
\label{tab:rh_milp_sensitivity}
\footnotesize
\setlength{\tabcolsep}{3.0pt}
\renewcommand{\arraystretch}{1.08}
\begin{tabular}{ccccc}
\hline
\multirow{2}{*}{\shortstack{Receding-horizon\\length, $H_{\mathrm{RH}}$}}
& \multicolumn{1}{c}{Time limit}
& \multicolumn{1}{c}{Relative}
& \multicolumn{1}{c}{Cumulative}
& \multicolumn{1}{c}{Planning time} \\
& \multicolumn{1}{c}{(s)}
& \multicolumn{1}{c}{MIP gap}
& \multicolumn{1}{c}{reward}
& \multicolumn{1}{c}{(ms)} \\
\hline
18 & 0.25 & $1\%$   & 10534.8 & 2.17 \\
36 & 0.25 & $1\%$   & 10690.2 & 4.79 \\
72 & 0.25 & $1\%$   & 10738.5 & 11.60 \\
36 & 1.00 & $1\%$   & 10722.8 & 6.35 \\
36 & 0.25 & $0.1\%$ & 10701.4 & 4.68 \\
\hline
\end{tabular}
\end{table}

As shown in \tabref{tab:rh_milp_sensitivity}, increasing the receding-horizon length from 18 to 36 decision steps improves the mean 10-day cumulative mission reward, indicating that a moderate planning look-ahead is beneficial for coordinating future observation opportunities, downlink windows, and resource evolution. Further extending the receding-horizon length to 72 steps yields only a modest additional reward improvement, whereas the mean planning time increases substantially from $4.79$ ms to $11.60$ ms. Increasing the solver time limit from $0.25$ s to $1.00$ s provides a small reward improvement at additional online planning cost. Likewise, tightening the relative MIP gap from $1\%$ to $0.1\%$ produces only a marginal reward increase. These results indicate that the nominal RH-MILP configuration captures most of the attainable benefit of a longer or more accurately solved finite-horizon optimization problem, while avoiding the markedly higher online planning burden associated with a substantially longer receding horizon.

Overall, the sensitivity analysis shows that the nominal reward setting provides a balanced trade-off among observation return, data downlink, and resource safety, while the physics-constrained action mask enforces action selection within the modeled admissible action set under strongly constrained mission states. In addition, the RH-MILP configuration analysis indicates that extending the receding horizon or increasing solver effort provides only marginal reward improvements while potentially increasing the online planning burden.

\subsection{Planning adaptability under uncertain conditions}\label{4.5}

In practical Titan autonomous mission planning, various types of uncertainty may arise during mission execution. These uncertainties can be broadly divided into internal resource disturbances, including energy degradation and onboard storage pressure, and external mission-environment disturbances, such as reduced communication opportunities and changes in target priority. To evaluate the adaptability and robustness of PCM-PPO under such uncertain conditions, five representative scenarios are designed, corresponding to the nominal condition, internal energy budget degradation, external communication degradation, coupled energy-communication degradation, and external target-priority variation. In all scenarios, the daily target load is fixed at 30, and each method is evaluated over a 10-day planning horizon. All learning-based policies are trained under the nominal setting and directly evaluated under the disturbed scenarios without retraining. Within each Monte Carlo trial, all methods are evaluated using the same sampled target realization and environmental random seed to ensure a fair comparison.

\begin{figure}[!t]
  \centering
  \includegraphics[width=7.5cm]{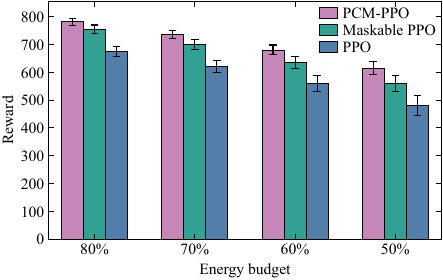}
  \caption{Average daily reward comparison under energy budget degradation in Scenario 2.}
  \label{fig4_12}
\end{figure}

\begin{figure}[!t]
  \centering
  \includegraphics[width=7.5cm]{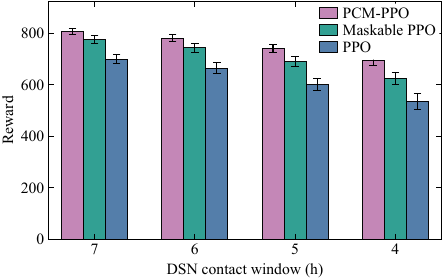}
  \caption{Average daily reward comparison under communication degradation in Scenario 3.}
  \label{fig4_13}
\end{figure}

\begin{figure}[!t]
  \centering
  \includegraphics[width=7.5cm]{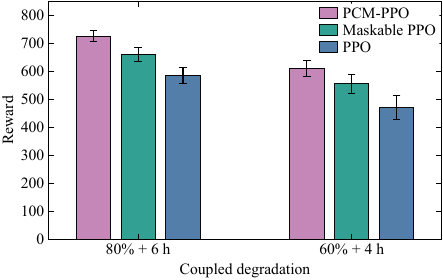}
  \caption{Average daily reward comparison under coupled energy-communication degradation in Scenario 4.}
  \label{fig4_14}
\end{figure}

For the internal energy budget degradation, external communication degradation, and coupled energy-communication degradation scenarios, the results are aggregated over 20 independent Monte Carlo trials. \figref{fig4_12}, \figref{fig4_13} and \figref{fig4_14} report the mean daily reward and the corresponding 95\% confidence intervals. The accompanying target-level tables provide illustrative scheduling outcomes for fixed realizations of the considered scenarios. Because target identities, arrival histories, and environmental realizations vary across trials, these tables are used to explain the scheduling behavior underlying the aggregate results rather than to summarize cross-trial performance.

\textit{Scenario 1: Nominal condition.}

\begin{table}[!t]
\centering
\caption{Illustrative observation results of PCM-PPO, Maskable PPO, and PPO under Scenario 1.}
\label{tab:scenario1_results}
\renewcommand{\arraystretch}{1.08}
\setlength{\tabcolsep}{4pt}
\footnotesize
\resizebox{\columnwidth}{!}{
\begin{tabular}{cccccc}
\hline
Task ID
& Maneuver demand
& Reward level
& PCM-PPO
& Maskable PPO
& PPO \\
\hline
4  & Low    & Opportunity & Success & Success & Success \\
9  & Low    & Opportunity & Success & Success & Success \\
27 & Low    & Opportunity & Success & Success & Success \\
11 & Medium & Opportunity & Success & Success & Success \\
5  & Low    & High        & Success & Success & Success \\
 10 & Low    & High        & Success& Success& Success\\
14 & Low    & High        & Success & Success & Success \\
21 & Low    & High        & Success & Success & Success \\
25 & Medium & High        & Success & Success & Success \\
2  & Medium & Low         & Success & Success & Success \\
3  & Low    & Low         & Success & Success & Success \\
7  & Low    & Low         & Success & Success & Success \\
12 & Low    & Low         & Success & Success & Success \\
16 & Medium & Low         & Success & Success & Success \\
18 & Low    & Low         & Success & Success & Success \\
24 & Low    & Low         & Success & Success & Success \\
29 & Low    & Low         & Success & Success & Success \\
\hline
8  & Medium & Opportunity & Success & Fail    & Fail \\
20 & Medium & High        & Success & Success & Fail \\
30 & High   & High        & Success & Success & Fail \\
\hline
26 & High   & Opportunity & Fail    & Fail    & Fail \\
23 & Medium & High        & Fail    & Fail    & Fail \\
28 & Low    & High        & Fail    & Fail    & Fail \\
1  & Low    & Low         & Fail    & Fail    & Fail \\
6  & Low    & Low         & Fail    & Fail    & Fail \\
13 & Low    & Low         & Fail    & Fail    & Fail \\
15 & Medium & Low         & Fail    & Fail    & Fail \\
17 & High   & Low         & Fail    & Fail    & Fail \\
19 & High   & Low         & Fail    & Fail    & Fail \\
22 & Low    & Low         & Fail    & Fail    & Fail \\
\hline
\end{tabular}
}
\end{table}

In this scenario, the probe operates under the standard resource and communication settings. As shown in \tabref{tab:scenario1_results}, all three methods complete most feasible observations, indicating that the nominal scenario provides a relatively moderate planning condition. PCM-PPO successfully completes 20 targets, while Maskable PPO and PPO complete 19 and 17 targets, respectively. The methods with the physics-constrained action mask achieve higher completion levels than PPO, suggesting that explicit feasibility filtering helps reduce ineffective action selection under the same mission state. In addition, PCM-PPO completes one more opportunity target than Maskable PPO, indicating that the target-pool representation and target-conditioned policy modeling further improve opportunity target prioritization beyond action masking.

\textit{Scenario 2: Internal energy budget degradation.}

\begin{table}[!t]
\centering
\caption{Illustrative observation results under Scenario 2 with an 80\% energy budget.}
\label{tab:scenario2_80_results}
\renewcommand{\arraystretch}{1.08}
\setlength{\tabcolsep}{4pt}
\footnotesize
\resizebox{\columnwidth}{!}{
\begin{tabular}{cccccc}
\hline
Task ID
& Maneuver demand
& Reward level
& PCM-PPO
& Maskable PPO
& PPO \\
\hline
9  & Low    & Opportunity & Success & Success & Success \\
27 & Low    & Opportunity & Success & Success & Success \\
11 & Medium & Opportunity & Success & Success & Success \\
5  & Low    & High        & Success & Success & Success \\
10 & Low    & High        & Success & Success & Success \\
14 & Low    & High        & Success & Success & Success \\
21 & Low    & High        & Success & Success & Success \\
 28 & Low    & High        & Success& Success&Success\\
25 & Medium & High        & Success & Success & Success \\
30 & High   & High        & Success & Success & Success \\
3  & Low    & Low         & Success & Success & Success \\
7  & Low    & Low         & Success & Success & Success \\
12 & Low    & Low         & Success & Success & Success \\
24& Low& Low         & Success & Success & Success \\
16& Medium& Low         & Success & Success & Success \\
\hline
4  & Low    & Opportunity & Success & Success & Fail \\
20 & Medium & High        & Success & Success & Fail \\
8  & Medium & Opportunity & Success & Fail    & Fail \\
\hline
26 & High   & Opportunity & Fail    & Fail    & Fail \\
23 & Medium & High        & Fail    & Fail    & Fail \\
1  & Low    & Low         & Fail    & Fail    & Fail \\
2  & Medium & Low         & Fail    & Fail    & Fail \\
6  & Low    & Low         & Fail    & Fail    & Fail \\
13 & Low    & Low         & Fail    & Fail    & Fail \\
15 & Medium & Low         & Fail    & Fail    & Fail \\
17 & High   & Low         & Fail    & Fail    & Fail \\
18 & Low    & Low         & Fail    & Fail    & Fail \\
19 & High   & Low         & Fail    & Fail    & Fail \\
22 & Low    & Low         & Fail    & Fail    & Fail \\
29 & Low    & Low         & Fail    & Fail    & Fail \\
\hline
\end{tabular}
}
\end{table}

\begin{table}[!t]
\centering
\caption{Illustrative observation results under Scenario 2 with a 60\% energy budget.}
\label{tab:scenario2_60_results}
\renewcommand{\arraystretch}{1.08}
\setlength{\tabcolsep}{4pt}
\footnotesize
\resizebox{\columnwidth}{!}{
\begin{tabular}{cccccc}
\hline
Task ID
& Maneuver demand
& Reward level
& PCM-PPO
& Maskable PPO
& PPO \\
\hline
9  & Low    & Opportunity & Success & Success & Success \\
27 & Low    & Opportunity & Success & Success & Success \\
5  & Low    & High        & Success & Success & Success \\
10 & Low    & High        & Success & Success & Success \\
14 & Low    & High        & Success & Success & Success \\
21 & Low    & High        & Success & Success & Success \\
 28 & Low    & High        & Success& Success&Success\\
25 & Medium & High        & Success & Success & Success \\
3  & Low    & Low         & Success & Success & Success \\
 12& Low    & Low         & Success& Success&Success\\
 22 & Low    & Low         & Success& Success&Success\\
\hline
 4  & Low    & Opportunity & Success & Success&Fail \\
 1& Low    & Low         & Success & Success&Fail \\
20 & Medium & High        & Success & Fail& Fail \\
11 & Medium & Opportunity & Success & Fail    & Fail \\
\hline
7& Low    & Low         & Fail    & Success & Success\\
\hline
8  & Medium & Opportunity & Fail    & Fail    & Fail \\
26 & High   & Opportunity & Fail    & Fail    & Fail \\
30 & High   & High        & Fail    & Fail    & Fail \\
23 & Medium & High        & Fail    & Fail    & Fail \\
2  & Medium & Low         & Fail    & Fail    & Fail \\
6  & Low    & Low         & Fail    & Fail    & Fail \\
13 & Low    & Low         & Fail    & Fail    & Fail \\
15 & Medium & Low         & Fail    & Fail    & Fail \\
16 & Medium & Low         & Fail    & Fail    & Fail \\
17 & High   & Low         & Fail    & Fail    & Fail \\
18 & Low    & Low         & Fail    & Fail    & Fail \\
19 & High   & Low         & Fail    & Fail    & Fail \\
24 & Low    & Low         & Fail    & Fail    & Fail \\
29 & Low    & Low         & Fail    & Fail    & Fail \\
\hline
\end{tabular}
}
\end{table}

In this scenario, an internal resource disturbance is introduced by reducing the available onboard energy budget to 80\%, 70\%, 60\%, and 50\% of the nominal setting. As shown in \figref{fig4_12}, the average daily rewards of all three methods decrease as the energy budget becomes more restrictive, indicating that energy-budget degradation directly limits feasible observation and downlink decisions. Nevertheless, PCM-PPO consistently obtains the highest reward under all energy-budget levels, followed by Maskable PPO and PPO. Meanwhile, the results in \tabref{tab:scenario2_80_results} and \tabref{tab:scenario2_60_results} further show this trend. With an 80\% energy budget, PCM-PPO completes 18 targets, while Maskable PPO and PPO complete 17 and 15 targets, respectively. When the energy budget is further reduced to 60\%, the completed targets decrease to 15, 14, and 12, respectively. These results indicate that all methods reduce observation attempts under tighter energy constraints, but PCM-PPO better preserves high-value and opportunity observations. This is because the physics-constrained action mask restricts actions that violate resource-safety requirements, while the target-pool representation and target-conditioned policy modeling help the policy allocate the remaining feasible observation opportunities toward targets with higher mission value.

\textit{Scenario 3: External communication degradation.}

\begin{table}[!t]
\centering
\caption{Illustrative observation results under Scenario 3 with a 6 h DSN contact window.}
\label{tab:scenario3_dsn6h_results}
\renewcommand{\arraystretch}{1.08}
\setlength{\tabcolsep}{4pt}
\footnotesize
\resizebox{\columnwidth}{!}{
\begin{tabular}{cccccc}
\hline
Task ID
& Maneuver demand
& Reward level
& PCM-PPO
& Maskable PPO
& PPO \\
\hline
9  & Low    & Opportunity & Success & Success & Success \\
27 & Low    & Opportunity & Success & Success & Success \\
11 & Medium & Opportunity & Success & Success & Success \\
5  & Low    & High        & Success & Success & Success \\
10 & Low    & High        & Success & Success & Success \\
14 & Low    & High        & Success & Success & Success \\
21 & Low    & High        & Success & Success & Success \\
25 & Medium & High        & Success & Success & Success \\
30 & High   & High        & Success & Success & Success \\
3  & Low    & Low         & Success & Success & Success \\
7  & Low    & Low         & Success & Success & Success \\
12 & Low    & Low         & Success & Success & Success \\
24 & Low    & Low         & Success & Success & Success \\
29 & Low    & Low         & Success & Success & Success \\
16 & Medium & Low         & Success & Success & Success \\
\hline
4  & Low    & Opportunity & Success & Success & Fail \\
8  & Medium & Opportunity & Success & Fail    & Fail \\
20 & Medium & High        & Success & Success & Fail \\
\hline
26 & High   & Opportunity & Fail    & Fail    & Fail \\
23 & Medium & High        & Fail    & Fail    & Fail \\
28 & Low    & High        & Fail    & Fail    & Fail \\
1  & Low    & Low         & Fail    & Fail    & Fail \\
2  & Medium & Low         & Fail    & Fail    & Fail \\
6  & Low    & Low         & Fail    & Fail    & Fail \\
13 & Low    & Low         & Fail    & Fail    & Fail \\
15 & Medium & Low         & Fail    & Fail    & Fail \\
17 & High   & Low         & Fail    & Fail    & Fail \\
18 & Low    & Low         & Fail    & Fail    & Fail \\
19 & High   & Low         & Fail    & Fail    & Fail \\
22 & Low    & Low         & Fail    & Fail    & Fail \\
\hline
\end{tabular}
}
\end{table}

In this scenario, an external communication disturbance is introduced by shortening the DSN contact window from the nominal 8 h setting to 7 h, 6 h, 5 h, and 4 h. Different from energy budget degradation, this disturbance does not directly reduce observation capability, but limits data downlink opportunities and increases onboard storage pressure, which may further constrain subsequent observation decisions. As shown in \figref{fig4_13}, the average daily rewards of all three methods decrease as the DSN contact window becomes shorter. PCM-PPO consistently achieves the highest reward under all communication-window settings, indicating that it can better coordinate observation selection, data downlink, and storage evolution under reduced communication availability. The results in \tabref{tab:scenario3_dsn6h_results} and \tabref{tab:scenario3_dsn4h_results} provide further evidence. With a 6 h DSN contact window, PCM-PPO completes 18 targets, while Maskable PPO and PPO complete 17 and 15 targets, respectively. When the contact window is shortened to 4 h, the completed targets decrease to 16, 13, and 11, respectively. These results show that reduced communication opportunities make all methods more conservative in selecting observations, while PCM-PPO still preserves more high-value and opportunity observations. This advantage is attributed to the communication availability and onboard storage states in the global mission state vector, together with the physics-constrained action mask and target-conditioned policy modeling, which enable PCM-PPO to select valuable feasible observations while avoiding decisions that would aggravate storage pressure.

\begin{table}[!t]
\centering
\caption{Illustrative observation results under Scenario 3 with a 4 h DSN contact window.}
\label{tab:scenario3_dsn4h_results}
\renewcommand{\arraystretch}{1.08}
\setlength{\tabcolsep}{4pt}
\footnotesize
\resizebox{\columnwidth}{!}{
\begin{tabular}{cccccc}
\hline
Task ID
& Maneuver demand
& Reward level
& PCM-PPO
& Maskable PPO
& PPO \\
\hline
9  & Low    & Opportunity & Success & Success & Success \\
27 & Low    & Opportunity & Success & Success & Success \\
5  & Low    & High        & Success & Success & Success \\
10 & Low    & High        & Success & Success & Success \\
14 & Low    & High        & Success & Success & Success \\
21 & Low    & High        & Success & Success & Success \\
25 & Medium & High        & Success & Success & Success \\
3  & Low    & Low         & Success & Success & Success \\
7  & Low    & Low         & Success & Success & Success \\
12 & Low    & Low         & Success & Success & Success \\
24 & Low    & Low         & Success & Success & Success \\
\hline
11 & Medium & Opportunity & Success & Success & Fail \\
4  & Low    & Opportunity & Success & Fail    & Fail \\
20 & Medium & High        & Success & Success & Fail \\
30 & High   & High        & Success & Fail    & Fail \\
16 & Medium & Low         & Success & Fail    & Fail \\
\hline
8  & Medium & Opportunity & Fail    & Fail    & Fail \\
26 & High   & Opportunity & Fail    & Fail    & Fail \\
23 & Medium & High        & Fail    & Fail    & Fail \\
28 & Low    & High        & Fail    & Fail    & Fail \\
1  & Low    & Low         & Fail    & Fail    & Fail \\
2  & Medium & Low         & Fail    & Fail    & Fail \\
6  & Low    & Low         & Fail    & Fail    & Fail \\
13 & Low    & Low         & Fail    & Fail    & Fail \\
15 & Medium & Low         & Fail    & Fail    & Fail \\
17 & High   & Low         & Fail    & Fail    & Fail \\
18 & Low    & Low         & Fail    & Fail    & Fail \\
19 & High   & Low         & Fail    & Fail    & Fail \\
22 & Low    & Low         & Fail    & Fail    & Fail \\
29 & Low    & Low         & Fail    & Fail    & Fail \\
\hline
\end{tabular}
}
\end{table}

\textit{Scenario 4: Coupled energy-communication degradation.}

\begin{table}[!t]
\centering
\caption{Illustrative observation results under Scenario 4 with an 80\% energy budget and a 6 h DSN contact window.}
\label{tab:scenario4_80e_6h}
\renewcommand{\arraystretch}{1.08}
\setlength{\tabcolsep}{4pt}
\footnotesize
\resizebox{\columnwidth}{!}{
\begin{tabular}{cccccc}
\hline
Task ID
& Maneuver demand
& Reward level
& PCM-PPO
& Maskable PPO
& PPO \\
\hline
9  & Low    & Opportunity & Success & Success & Success \\
27 & Low    & Opportunity & Success & Success & Success \\
11 & Medium & Opportunity & Success & Success & Success \\
5  & Low    & High        & Success & Success & Success \\
10 & Low    & High        & Success & Success & Success \\
14 & Low    & High        & Success & Success & Success \\
21 & Low    & High        & Success & Success & Success \\
25 & Medium & High        & Success & Success & Success \\
3  & Low    & Low         & Success & Success & Success \\
7  & Low    & Low         & Success & Success & Success \\
12 & Low    & Low         & Success & Success & Success \\
24 & Low    & Low         & Success & Success & Success \\
\hline
4  & Low    & Opportunity & Success & Success & Fail \\
8  & Medium & Opportunity & Success & Fail    & Fail \\
20 & Medium & High        & Success & Success & Fail \\
16 & Medium & Low         & Success & Fail    & Fail \\
\hline
26 & High   & Opportunity & Fail    & Fail    & Fail \\
23 & Medium & High        & Fail    & Fail    & Fail \\
28 & Low    & High        & Fail    & Fail    & Fail \\
30 & High   & High        & Fail    & Fail    & Fail \\
1  & Low    & Low         & Fail    & Fail    & Fail \\
2  & Medium & Low         & Fail    & Fail    & Fail \\
6  & Low    & Low         & Fail    & Fail    & Fail \\
13 & Low    & Low         & Fail    & Fail    & Fail \\
15 & Medium & Low         & Fail    & Fail    & Fail \\
17 & High   & Low         & Fail    & Fail    & Fail \\
18 & Low    & Low         & Fail    & Fail    & Fail \\
19 & High   & Low         & Fail    & Fail    & Fail \\
22 & Low    & Low         & Fail    & Fail    & Fail \\
29 & Low    & Low         & Fail    & Fail    & Fail \\
\hline
\end{tabular}
}
\end{table}

\begin{table}[!t]
\centering
\caption{Illustrative observation results under Scenario 4 with a 60\% energy budget and a 4 h DSN contact window.}
\label{tab:scenario4-60e4h}
\renewcommand{\arraystretch}{1.08}
\setlength{\tabcolsep}{4pt}
\footnotesize
\resizebox{\columnwidth}{!}{
\begin{tabular}{cccccc}
\hline
Task ID
& Maneuver demand
& Reward level
& PCM-PPO
& Maskable PPO
& PPO \\
\hline
9  & Low    & Opportunity & Success & Success & Success \\
27 & Low    & Opportunity & Success & Success & Success \\
5  & Low    & High        & Success & Success & Success \\
10 & Low    & High        & Success & Success & Success \\
14 & Low    & High        & Success & Success & Success \\
21 & Low    & High        & Success & Success & Success \\
25 & Medium & High        & Success & Success & Success \\
3  & Low    & Low         & Success & Success & Success \\
12 & Low    & Low         & Success & Success & Success \\
\hline
4  & Low    & Opportunity & Success & Success & Fail \\
11 & Medium & Opportunity & Success & Fail    & Fail \\
20 & Medium & High        & Success & Success & Fail \\
\hline
8  & Medium & Opportunity & Fail    & Fail    & Fail \\
26 & High   & Opportunity & Fail    & Fail    & Fail \\
30 & High   & High        & Fail    & Fail    & Fail \\
28 & Low    & High        & Fail    & Fail    & Fail \\
1  & Low    & Low         & Fail    & Fail    & Fail \\
2  & Medium & Low         & Fail    & Fail    & Fail \\
6  & Low    & Low         & Fail    & Fail    & Fail \\
7  & Low    & Low         & Fail    & Fail    & Fail \\
13 & Low    & Low         & Fail    & Fail    & Fail \\
15 & Medium & Low         & Fail    & Fail    & Fail \\
16 & Medium & Low         & Fail    & Fail    & Fail \\
17 & High   & Low         & Fail    & Fail    & Fail \\
18 & Low    & Low         & Fail    & Fail    & Fail \\
19 & High   & Low         & Fail    & Fail    & Fail \\
22 & Low    & Low         & Fail    & Fail    & Fail \\
23 & Medium & High        & Fail    & Fail    & Fail \\
24 & Low    & Low         & Fail    & Fail    & Fail \\
29 & Low    & Low         & Fail    & Fail    & Fail \\
\hline
\end{tabular}
}
\end{table}

In this scenario, internal resource disturbance and external communication disturbance are introduced simultaneously to evaluate planning adaptability under coupled uncertainty. Two degradation levels are considered: a moderate coupled degradation setting with an 80\% energy budget and a 6 h DSN contact window, and a severe coupled degradation setting with a 60\% energy budget and a 4 h DSN contact window. As shown in \figref{fig4_14}, the average daily rewards of all three methods decrease when the coupled degradation becomes more severe, indicating that simultaneous energy-budget reduction and communication-window shortening impose stronger constraints on autonomous planning than a single disturbance. Nevertheless, PCM-PPO maintains the highest reward under both coupled degradation settings. As shown in \tabref{tab:scenario4_80e_6h} and \tabref{tab:scenario4-60e4h}, under the moderate coupled degradation setting, PCM-PPO completes 16 targets, while Maskable PPO and PPO complete 14 and 12 targets, respectively. Under the severe coupled degradation setting, the completed targets decrease to 12, 11, and 9, respectively. These results indicate that the coupled disturbance reduces both the feasible observation capacity and the ability to relieve onboard storage pressure through downlink. Compared with the other two methods, PCM-PPO better preserves high-value and opportunity observations under the remaining feasible decision space. This suggests that the combination of global mission-state representation, physics-constrained action masking, and target-conditioned policy modeling improves planning adaptability when energy and communication constraints become simultaneously tighter.

\textit{Scenario 5: External target-priority variation.}

\begin{table}[!t]
\centering
\caption{Monte Carlo aggregate results under Scenario 5.}
\label{tab:priority_variation_mc}
\footnotesize
\setlength{\tabcolsep}{3pt}
\renewcommand{\arraystretch}{1.10}
\begin{tabularx}{\columnwidth}{@{}>{\centering\arraybackslash}p{0.25\columnwidth} >{\centering\arraybackslash}p{0.32\columnwidth} >{\centering\arraybackslash}X@{}}
\hline
\raisebox{0.55\baselineskip}{Method}
&
\shortstack[c]{Mean daily reward\\(95\% CI)}
&
\shortstack[c]{Mean opportunity-target\\completion rate}
\\
\hline
PCM-PPO      & $872.7 \pm 12.4$ & $0.771$ \\
Maskable PPO & $751.1 \pm 14.9$ & $0.574$ \\
PPO          & $726.1 \pm 16.8$ & $0.548$ \\
\hline
\end{tabularx}
\end{table}

\begin{table}[!t]
\centering
\caption{Illustrative observation results under Scenario 5 with target-priority variation.}
\label{tab:scenario5}
\renewcommand{\arraystretch}{1.08}
\setlength{\tabcolsep}{4pt}
\footnotesize
\resizebox{\columnwidth}{!}{
\begin{tabular}{cccccc}
\hline
Task ID
& Maneuver demand
& Reward level
& PCM-PPO
& Maskable PPO
& PPO \\
\hline
9  & Low    & Opportunity & Success & Success & Success \\
27 & Low    & Opportunity & Success & Success & Success \\
11 & Medium & Opportunity & Success & Success & Success \\
2  & Medium & Opportunity & Success & Success & Success \\
18 & Low    & Opportunity & Success & Success & Success \\
5  & Low    & High        & Success & Success & Success \\
10 & Low    & High        & Success & Success & Success \\
14 & Low    & High        & Success & Success & Success \\
21 & Low    & High        & Success & Success & Success \\
25 & Medium & High        & Success & Success & Success \\
1  & Low    & Low         & Success    & Success    & Success \\
3  & Low    & Low         & Success & Success & Success \\
7  & Low    & Low         & Success & Success & Success \\
12 & Low    & Low         & Success & Success & Success \\
16 & Medium & Low         & Success  & Success & Success \\
22 & Low    & Low         & Success  & Success  & Success \\
\hline
4  & Low    & Opportunity & Success & Fail    & Fail \\
24 & Low    & Opportunity & Success & Fail    & Fail \\
20 & Medium & High        & Success & Success & Fail \\
30 & High   & High        & Success & Success    & Fail \\
\hline
6  & Low    & Low         & Fail    & Success    & Success \\
\hline
8  & Medium & Opportunity & Fail    & Fail    & Fail \\
26 & High   & Opportunity & Fail    & Fail    & Fail \\
28 & Low    & High        & Fail    & Fail    & Fail \\
13 & Low    & Low         & Fail    & Fail    & Fail \\
15 & Medium & Low         & Fail    & Fail    & Fail \\
17 & High   & Low         & Fail    & Fail    & Fail \\
19 & High   & Low         & Fail    & Fail    & Fail \\
23 & Medium & High        & Fail    & Fail    & Fail \\
29 & Low    & Low         & Fail    & Fail    & Fail \\
\hline
\end{tabular}
}
\end{table}

In this scenario, an external mission disturbance is introduced to represent a change in the target-priority distribution. Specifically, a subset of the low-value targets is upgraded to opportunity targets, and the target composition is changed from 50\% low-value, 30\% high-value, and 20\% opportunity targets to 40\%, 30\%, and 30\%, respectively. This setting evaluates whether the policy can adjust its observation preference when the relative mission value and time sensitivity of candidate targets change.

To quantify the aggregate performance under target-priority variation, \tabref{tab:priority_variation_mc} reports the mean daily reward with its 95\% confidence interval and the mean opportunity-target completion rate over 20 independent Monte Carlo trials. As summarized in \tabref{tab:priority_variation_mc}, PCM-PPO achieves the highest mean daily reward of $872.7$, compared with $751.1$ for Maskable PPO and $726.1$ for PPO. The corresponding mean opportunity-target completion rate of PCM-PPO is $0.771$, which is higher than the corresponding rates of Maskable PPO ($0.574$) and PPO ($0.548$). Compared with the nominal target composition, the increased proportion of opportunity targets raises the potential available mission value. PCM-PPO is able to convert this additional value into a larger reward improvement by preferentially completing more time-sensitive opportunity targets, whereas Maskable PPO and PPO obtain smaller reward improvements.

In the illustrative realization reported in \tabref{tab:scenario5}, PCM-PPO completes 20 targets, while Maskable PPO and PPO complete 19 and 17 targets, respectively. More importantly, PCM-PPO completes 7 opportunity targets, whereas Maskable PPO and PPO each complete 5 opportunity targets. Meanwhile, PCM-PPO completes fewer low-value targets than the other two methods, indicating that it reallocates observation opportunities toward more time-sensitive and higher-priority targets rather than simply increasing the number of completed observations or relying on easier low-value targets. These results suggest that the target-pool representation and target-conditioned policy modeling help PCM-PPO perceive the changed target-value distribution and adjust target-level prioritization under dynamic mission priorities.

In summary, the uncertainty experiments show that PCM-PPO maintains favorable planning adaptability under different disturbed mission conditions, including internal energy budget degradation, external communication degradation, coupled energy-communication degradation, and external target-priority variation. In the scenarios with reported reward comparisons, PCM-PPO achieves the highest average daily reward, and it also obtains better target observation outcomes across all evaluated scenarios. Under resource and communication disturbances, PCM-PPO better preserves high-value and opportunity observations while reducing ineffective decisions caused by tightened energy, storage, and downlink constraints. Under target-priority variation, PCM-PPO further shows a stronger ability to adjust target-level prioritization toward time-sensitive opportunity targets. These results indicate that the proposed target-pool representation, target-conditioned policy modeling, and physics-constrained action feasibility mask jointly improve the effectiveness of autonomous Titan mission planning under uncertain conditions.

\section{Conclusion and discussion}\label{5}
This study investigates the autonomous mission planning problem for Titan deep space probes under complex physical constraints, resource limitations, and uncertain mission conditions. A Titan-specific discrete-time MDP is formulated to describe observation and downlink feasibility, onboard resource evolution, time-varying target utility, and environmental uncertainty. Based on this formulation, a Physics-Constrained Maskable Proximal Policy Optimization (PCM-PPO) method is proposed. The method represents predefined and opportunity targets through a fixed-capacity target pool, organizes the mission state into a target-wise state matrix and a global mission-state vector, and generates action logits through a target-conditioned actor structure. A physics-constrained action mask is further embedded into the policy distribution so that action sampling and policy updating are restricted to the feasible decision domain determined by the current mission state.

Simulation results show that PCM-PPO achieves better planning performance than representative learning-based, optimization-based, and metaheuristic baselines under varying target loads. The proposed method also maintains stable performance over the 100-day evaluation horizon and exhibits lower sensitivity to training randomness among PPO-based methods. The ablation results verify the contributions of the fixed-capacity target-pool representation, target-conditioned policy modeling, physics-constrained action mask, and integrated training strategy, while the sensitivity analysis further shows that the nominal reward setting provides a balanced trade-off among observation return, data downlink, and resource safety. In addition, the uncertainty experiments demonstrate that PCM-PPO maintains higher average daily rewards and better target observation outcomes than Maskable PPO and PPO under energy budget degradation, communication degradation, coupled energy-communication degradation, and target-priority variation.

It should be noted that the present study is still based on a simulated Titan mission environment. Although the proposed framework incorporates representative physical constraints and uncertainty factors, its onboard deployability has not yet been verified under realistic flight software, computation, memory, and communication-delay constraints. In addition, the current experiments focus on single-probe autonomous planning, while cooperative planning among multiple deep space probes and more detailed operational constraints remain to be further investigated. Moreover, the dynamic addition of opportunity targets is supported only within the preallocated target-pool capacity; the current implementation does not consider an unbounded number of newly discovered targets or target admission and replacement when all reserved opportunity-target slots are occupied.

Future work will focus on improving the onboard implementation feasibility of PCM-PPO, including lightweight policy deployment, real-time inference under limited computational resources, and software-in-the-loop or hardware-in-the-loop validation. Further studies will also consider more realistic communication-delay effects, adaptive communication-resource management, and cooperative autonomous planning among multiple deep space probes.

\section*{Acknowledgements}
The study was supported by the Civil Space Technology Advance Research Project (No. D010103).
\nocite{*}
\bibliographystyle{cja}
\bibliography{references}
\end{document}